\documentclass[pre,twocolumn,showpacs,showkeys,superscriptaddress,preprintnumbers,floatfix,dvips]{revtex4}

\usepackage{newcent}
\usepackage{varioref}
\usepackage{oubraces}

\usepackage{array,graphicx,amsfonts,amsmath,amsthm}
\usepackage{url}
\makeatletter

\usepackage{amstext}
\usepackage{amssymb}
\usepackage{stmaryrd}
\DeclareFontFamily{OT1}{cmtex}{}
\DeclareFontShape{OT1}{cmtex}{m}{n}
  {<5><6><7><8>cmtex8
   <9>cmtex9
   <10><10.95><12><14.4><17.28><20.74><24.88>cmtex10}{}
\DeclareFontShape{OT1}{cmtex}{m}{it}
  {<-> ssub * cmtt/m/it}{}

\DeclareFontShape{OT1}{cmtt}{bx}{n}
  {<5><6><7><8>cmtt8
   <9>cmbtt9
   <10><10.95><12><14.4><17.28><20.74><24.88>cmbtt10}{}
\DeclareFontShape{OT1}{cmtex}{bx}{n}
  {<-> ssub * cmtt/bx/n}{}

\newlength{\lwidth}
\newlength{\cwidth}

\newcommand{\Conid}[1]{\mathit{#1}}
\newcommand{\Varid}[1]{\mathit{#1}}
\newcommand{\anonymous}{\kern0.06em \vbox{\hrule\@width.5em}}\newcommand{\plus}{\mathbin{+\!\!\!+}}

\newcommand{\sub}{\ensuremath{\mrm{sub}}}

\newcommand{\Figure}{Fig.}
\renewcommand{\vref}{\ref}

\makeatother

\usepackage[pdfauthor={Carl S. McTague and James
  P. Crutchfield},pdftitle={Automated Pattern
  Detection},bookmarks=true]{hyperref}

\newcommand{\Nn}{\ensuremath{\mathbb{N}}}
\newcommand{\Zr}{\ensuremath{\mathbb{Z}}}

\newcommand{\mc}[1]{\mathcal{#1}}
\newcommand{\mrm}[1]{\mathrm{#1}}

\newcommand{\Input}{\ensuremath{\mathrm{In}}}
\newcommand{\Det}{\ensuremath{\mathrm{Det}}}
\newcommand{\Filter}{\ensuremath{\mathrm{Filter}}}
\newcommand{\Optimize}{\ensuremath{\mathrm{Optimize}}}
\newcommand{\Output}{\ensuremath{\mathrm{Out}}}
\newcommand{\Lang}{\ensuremath{\mathrm{Lang}}}

\newcommand{\CA}{\textsc{ca}}
\newcommand{\CAs}{\textsc{ca}s}
\newcommand{\CapitalCAs}{C\textsc{a}s}
\newcommand{\CapitalCA}{C\textsc{a}}
\newcommand{\CapitalECA}{E\textsc{ca}}
\newcommand{\ECA}{\textsc{eca}}
\newcommand{\ECAs}{\textsc{eca}s}

\newtheorem{prop}{Proposition}
\newtheorem{lem}{Lemma}
\newtheorem{alg}{Algorithm}

\newcommand{\domain}[1]{\ensuremath{\mc{D}_{#1}}}

\newcommand{\trans}[3]{\ensuremath{(#1,#2,#3)}}

\begin{document}

\title[Automated Pattern Detection]{Automated Pattern Detection---\\
An Algorithm for Constructing Optimally\\
Synchronizing Multi-Regular Language Filters}

\author{Carl S. McTague}
\homepage{www.mctague.org/carl}
\email{mctague@santafe.edu}
\affiliation{Santa Fe Institute, 1399 Hyde Park Road, Santa Fe, NM 87501, USA}
\affiliation{Department of Mathematical Sciences, University of
  Cincinnati, Cincinnati, OH 45221-0025, USA}

\author{James P. Crutchfield}
\homepage{www.santafe.edu/~chaos}
\email{chaos@santafe.edu}
\affiliation{Santa Fe Institute, 1399 Hyde Park Road, Santa Fe, NM 87501, USA}

\date{\today}

\begin{abstract}
  In the computational-mechanics structural analysis of
  one-dimensional cellular automata the following automata-theoretic
  analogue of the \emph{change-point problem} from time series
  analysis arises: \emph{Given a string $\sigma$ and a collection
    $\{\mc{D}_i\}$ of finite automata, identify the regions of
    $\sigma$ that belong to each $\mc{D}_i$ and, in particular, the
    boundaries separating them.}  We present two methods for solving
  this \emph{multi-regular language filtering problem}.  The first,
  although providing the ideal solution, requires a stack, has a
  worst-case compute time that grows quadratically in $\sigma$'s
  length and conditions its output at any point on arbitrarily long
  windows of future input.  The second method is to algorithmically
  construct a transducer that approximates the first algorithm.  In
  contrast to the stack-based algorithm, however, the transducer
  requires only a finite amount of memory, runs in linear time, and
  gives immediate output for each letter read; it is, moreover, the
  best possible finite-state approximation with these three features.
\end{abstract}

\pacs{
  05.45.Tp, 
  89.70.+c, 
  05.45.-a, 
  89.75.Kd  
}

\keywords{cellular automata; regular languages; computational
  mechanics; domains; particles; pattern detection; transducer;
  filter; synchronization; change-point problem}

\preprint{Santa Fe Institute Working Paper 04-09-027}
\preprint{arxiv.org/abs/cs.CV/0409XXX}
%

\maketitle



\section{Introduction}

Imagine you are confronted with an immense one-dimensional dataset in
the form of a string $\sigma$ of letters from a finite alphabet
$\Sigma$.  Suppose moreover that you discover that vast expanses of
$\sigma$ are regular in the sense that they are recognized by simple
finite automata~$\mc{D}_1, \dots, \mc{D}_n$.  You might wish to bleach
out these regular substrings so that only the boundaries separating
them remain, for this reduced presentation might illuminate $\sigma$'s
more subtle, larger-scale structure.

This \emph{multi-regular language filtering problem} is the
automata-theoretic analogue of several, more statistical, problems
that arise in a wide range of disciplines.  Examples include
estimating stationary epochs within time series (known as the
\emph{change-point problem} \cite{Zack83a}), distinguishing gene
sequences and promoter regions from enveloping junk \textsc{dna}
\cite{Bald01a}, detecting phonemes in sampled speech \cite{Furu89a},
and identifying regular segments within line-drawings \cite{Free74a},
to mention a few.

The multi-regular language filtering problem arises directly in the
computational-mechanics structural analysis of cellular automata
\cite{Crut92c}.  There, finite automata recognizing temporally invariant
sets of strings are identified and then filtered from space-time diagrams
to reveal systems of particles whose interactions capture the essence of
how a cellular automaton processes spatially distributed information.

We present two methods for solving the multi-regular language
filtering problem.  The first covers $\sigma$ with maximal substrings
recognized by the automata~$\{\mc{D}_i\}$.  The interesting parts of
$\sigma$ are then located where these segments overlap or abut.
Although this approach provides the ideal solution to the problem, it
unfortunately requires an arbitrarily deep stack to compute, has a
worst-case compute time that grows quadratically in $\sigma$'s length,
and conditions its output at any point on arbitrarily long windows of
future input.  As a result, this method becomes extremely expensive to
compute for large data sets, including the expansive space-time
diagrams that researchers of cellular automata often scrutinize.

The second method---and our primary focus---is to
algorithmically construct a finite transducer that approximates the
first, stack-based algorithm by printing sequences of labels $i$ over
segments of $\sigma$ recognized by the automaton~$\mc{D}_i$.  When, at
the end of such a segment, the transducer encounters a letter
forbidden by the prevailing automaton~$\mc{D}_i$, it prints special
symbols until it resynchronizes to a new automaton $\mc{D}_j$.  In
this way, the transducer approximates the stack-based algorithm by
jumping from one maximal substring to the next, printing a few special
symbols in between.  Since it does not jump to a new maximal substring
until the preceding one ends, however, the transducer can miss the true
beginning of any maximal substring that overlaps with the preceding one.
Typically, the benefits of the finite transducer outweigh the occurrence
of such errors.

In contrast with the stack-based algorithm it approximates, however,
the transducer requires only a finite amount of memory, runs in linear
time, and gives immediate output for each letter read---significant
improvements for cellular automata structural analysis and, we
suspect, for other applications as well. Put more precisely, the
transducer is Lipschitz-continuous (with Lipschitz constant one) under
the cylinder-set topology, whereas the stack-based algorithm, which
conditions its output on arbitrarily long windows of future input, is
generally not even continuous.

It is also worth noting that the transducers thus produced are the
best possible approximations with these three features and are
identical to those that researchers have historically constructed by
hand.  Our algorithm thus relieves researchers of the tedium of
constructing ever more complicated transducers.

\subsection*{Cellular Automata}
\label{ca}

Before presenting our two filtering methods, we introduce cellular
automata in order to highlight an important setting where the
multi-regular language filtering problem arises, as well as to give
some visual intuition to our approach.

Let $\Sigma$ be a discrete alphabet of $k$~symbols.  A \emph{local
  update rule} of radius~$r$ is any function $\phi : \Sigma^{2r+1}
\rightarrow \Sigma$.  Given such a function, we can construct a global
mapping of bi-infinite strings $\Phi : \Sigma^\Zr \rightarrow
\Sigma^\Zr$, called a \emph{one-dimensional cellular automaton}~(\CA),
by setting:
\begin{align*}
  \Phi(\sigma)_i := \phi(\sigma_{i-r} \dots 
  \sigma_i \dots \sigma_{i+r}) ~,
\end{align*}
where $\sigma_i$ denotes the $i$th letter of the string $\sigma$.
Since the image under $\Phi$ of any period-$N$ bi-infinite string
also has period $N$, it is common to regard $\Phi$ as a mapping of
finite strings, $\Sigma^N \rightarrow \Sigma^N$.  When regarded in
this way, a \CA\ is said to have \emph{periodic boundary conditions}.

For $k$=2 and $r$=1, there are precisely 256 local update rules, and
the resulting \CAs\ are called the \emph{elementary} \CAs\ (or \ECAs).
Wolfram~\cite{Wolfram83} introduced a numbering scheme for them: Order
the neighborhoods $\Sigma^{3}$ lexicographically and interpret the
symbols $\{\phi(\eta) : \eta \in \Sigma^{3}\}$ as the binary
representation of an integer between 0 and 255.

\begin{figure}
\includegraphics[width=3in]{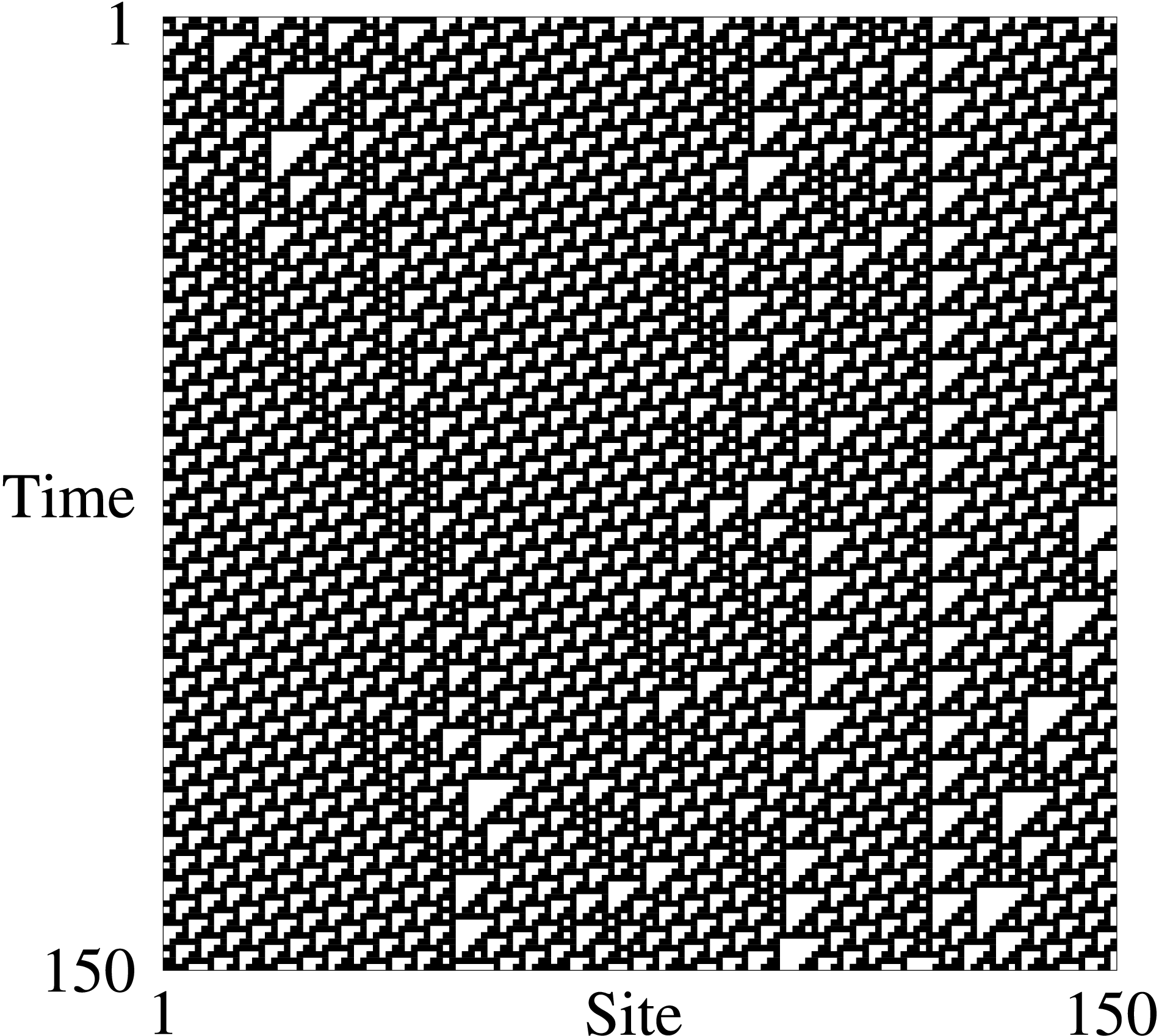}
\caption{A space-time diagram illustrating the typical behavior of
  \ECA~110. Black squares correspond to 1s, and white squares to 0s.
  \label{eca110}}
\end{figure}

By interpreting a string's letters as values assumed by the
\emph{sites} of a discrete \emph{lattice}, a \CA\ can be viewed as a
spatially extended dynamical system---discrete in time, space, and
local state.  Its behavior as such is often illustrated through
so-called \emph{space-time diagrams}, in which the iterates $\{
\Phi^t(\sigma^0) \}_{t = 0, 1, 2, \dots}$ of an initial string
$\sigma^0$ are plotted as a function of time.  Figure~\vref{eca110},
for example, depicts \ECA~110 acting iteratively on an initial string
of length $N=150$.

Due to their appealingly simple architecture, researchers have studied
\CAs\ not only as abstract mathematical objects, but as models for
physical, chemical, biological, and social phenomena such as fluid
flow, galaxy formation, earthquakes, chemical pattern formation,
biological morphogenesis, and vehicular traffic dynamics.
Additionally, they have been used as parallel computing devices, both
for the high-speed simulation of scientific models and for
computational tasks such as image processing.  More generally, \CAs\ 
have provided a simplified setting for studying the ``emergence'' of
cooperative or collective behavior in complex systems.  The literature
for all these applications is vast and includes Refs.~\cite{Burks70a,
  FarmerEtAL84a, FogelmanSoulieEtAl87, Gutowitz90a, Kara94a, Nage96a,
  PARCELLA94, Crut98c, Toffoli&Margolus87, Wolfram86a}.

\subsection*{Computational-Mechanics Structural Analysis of \CapitalCAs}

The computational-mechanics \cite{Crutchfield&Young89,Crut98d}
structural analysis of a \CA\ rests on the discovery of a ``pattern
basis''---a collection $\{\mc{D}_i\}$ of automata that describe the
emergent structural components in the \CA's space-time behavior
\cite{Crut93a,Hans90a}.  Once such a pattern basis is found, conforming
regions of space-time can be seen as background \emph{domains} through
which coherent structures not fitting the basis move. In this way,
structural features set against the domains can be identified and
analyzed.

\begin{figure*}
\includegraphics[scale=0.4]{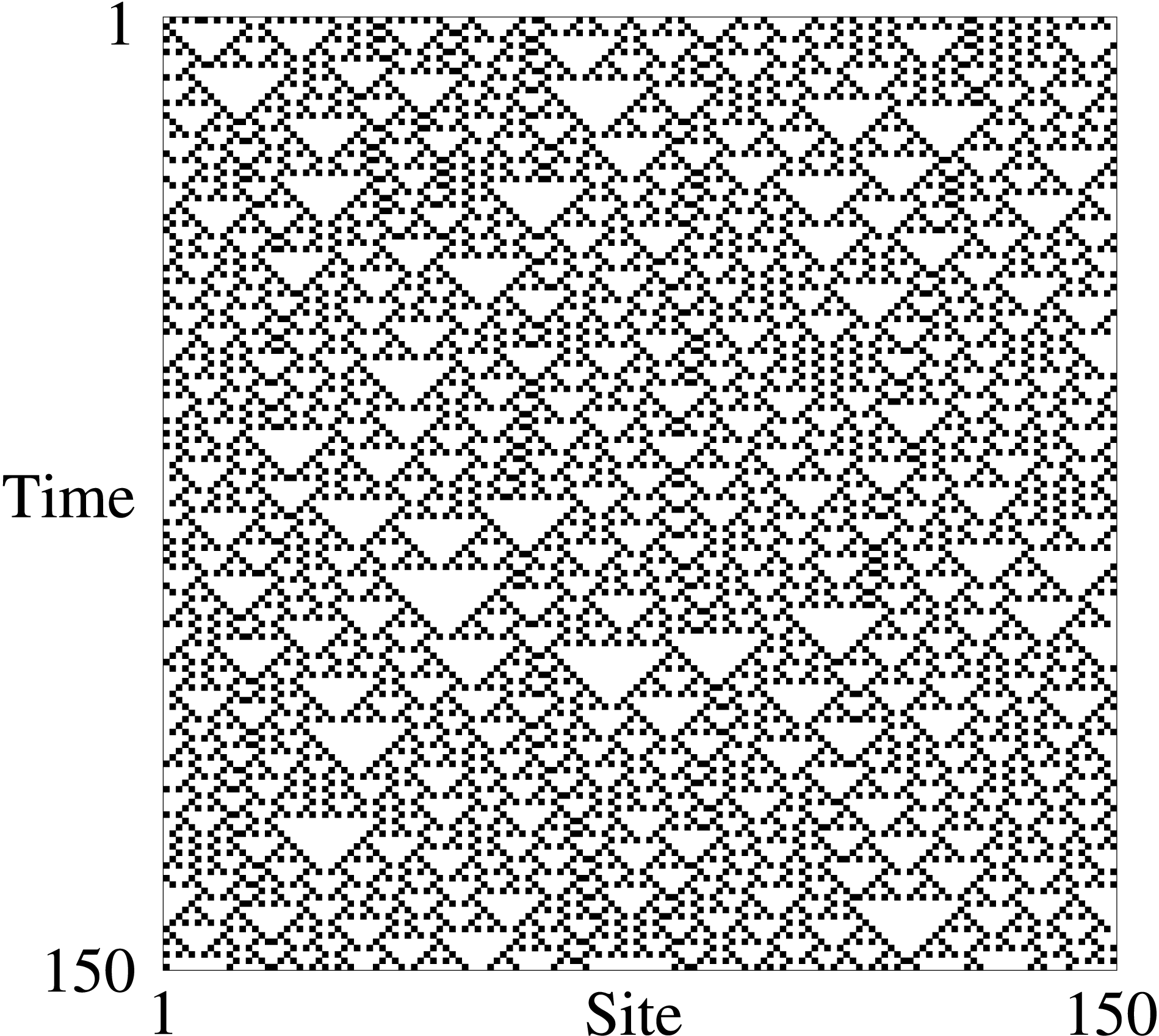}
\includegraphics[scale=0.4]{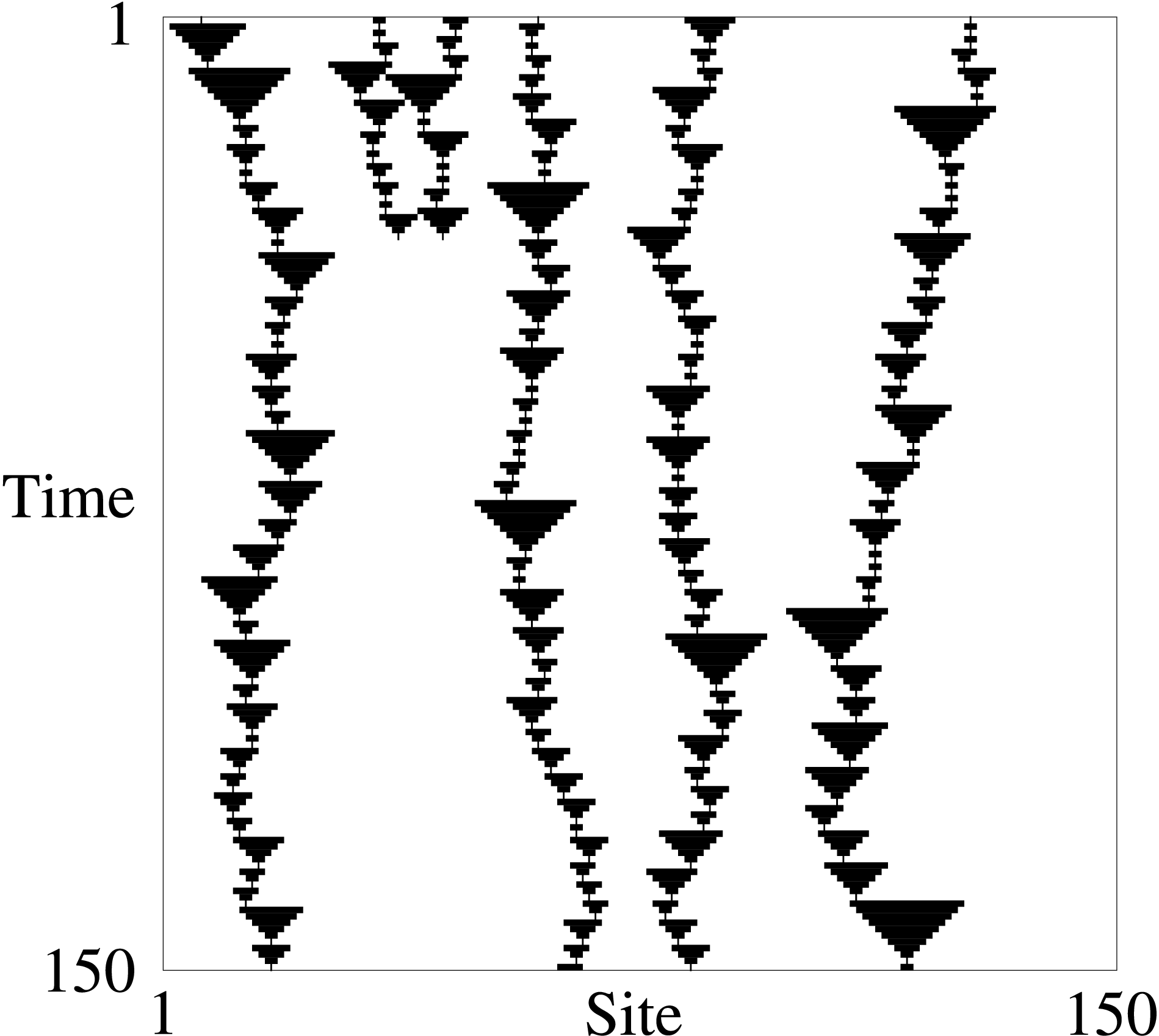}
\caption{(Left)~Space-time diagram illustrating the typical behavior of
  \ECA~18---a \CA\ exhibiting apparently random behavior, i.e., the
  set of length-$L$ spatial strings has a positive entropy density as
  $L \rightarrow \infty$.  (Right)~The same space-time diagram
  filtered with the regular domain $\mc{D}=\sub \left([0(0+1)]^*
  \right)$. (After Ref.~\cite{Crut92a}.)
\label{eca18}}
\end{figure*}

More formally, Crutchfield and Hanson define a \emph{regular domain}
\domain{} to be a regular language (the collection of strings
recognized by some finite automaton) that is:
\begin{enumerate}
\item \emph{temporally invariant}---the \CA\ maps \domain{}
  \emph{onto} itself; that is, $\Phi^n[\domain{}] = \domain{}$ for
  some $n>0$\ ---and
\item \emph{spatially homogeneous}---the same pattern can occur at any
  letter: the recurrent states in the minimal finite automaton
  recognizing $\domain{}$ are strongly connected.
\end{enumerate}

Once we discover a \CA's regular domains---either through visual
inspection or by an automated induction method such as the
$\epsilon$-machine reconstruction algorithm
\cite{Crut92c}---the corresponding space-time regions are, in
a sense, understood.  Given this level of discovered regularity, we
bleach out the domain-conforming regions from space-time diagrams,
leaving only ``un-modeled'' deviations, whose dynamics can then be
studied.  Sometimes, as is the case for the \CAs\ we exhibit here,
these deviations resemble particles and, by studying the
characteristics of these particle-like deviations---how they move and
what happens when they collide, we hope to understand the \CA's
(possibly hidden) computational capabilities.

Consider, for example, the apparently random behavior of \ECA~18,
illustrated in \Figure~\vref{eca18}.  Although no coherent structures
present themselves to the eye, computational-mechanics structural
analysis lays bare particles hidden within its output: Filtering
its space-time diagrams with the regular domain $\mc{D} =
\sub([0(0+1)]^*)$---where $\sub(\mc{L})$ denotes the regular language
consisting of all subwords of strings belonging to the regular
language $\mc{L}$---reveals a system of particles that follow random
walks and pairwise annihilate whenever they
touch~\cite{Crut92a,Eloranta94,Hans90a}.  Thus, by blurring the \CA's
deterministic behavior on strings, we discover higher-level stochastic
particle dynamics.  Although this loss of deterministic detail may at
first seem conceptually unsatisfying, the resulting view is more
structurally detailed than the vague classification of \ECA~18 as
``chaotic''.

Thus, discovering domains and filtering them from space-time diagrams
is essential to understanding the information processing embedded
within a \textsc{ca}'s output.

\section{Method 1---Filtering with a Stack}

We now present the first method for solving the general multi-regular
language filtering problem with which we began. Although
the following method is perhaps the most thorough and easiest to
describe, it requires an arbitrarily deep stack to compute.  Its
description will rest upon a few basic ideas from automata theory.
(Please refer to the first few paragraphs of App.~\vref{automata-review},
up to and including Lemma~\ref{subset-construction}, where these
preliminaries are reviewed.)

To filter a string $\sigma$, this method identifies the collection of
its maximal substrings that the automata $\{\mc{D}_i\}$ accept.  More
formally, given a string $\sigma$, let $\sigma_{a,b}$ denote the
substring $\sigma_a \sigma_{a+1} \cdots \sigma_b$ for $a, b \in \Zr$.
If $\sigma$ is bi-infinite, extend this notation so that $a=-\infty$
and $b=\infty$ denote the intuitive infinite substrings.  Place a
partial ordering $\prec$ on all such substrings by setting
$\sigma_{a,b} \prec \sigma_{a',b'}$ if $a' \le a \le b \le b'$.  Then
let $\mc{P}_\mrm{max}(\{\mc{D}_i\},\sigma)$ denote the collection of
maximal substrings $\sigma_{a,b}$ (with respect to $\prec$) that the
$\{\mc{D}_i\}$ accept---or, in symbols, let:
\begin{align*}
  \mc{P}_\mrm{max}(\{\mc{D}_i\},\sigma) := \{ \sigma_{a,b} \in \mc{P}
  :\ & \text{there is no $\sigma' \in \mc{P}$} \\ &\text{with
    $\sigma_{a,b} \prec \sigma'$} \} ~,
\end{align*}
where $\mc{P} := \{ \sigma_{a,b} : \text{$\mc{D}_i$ accepts
  $\sigma_{a,b}$ for some $i$}\}$.

The following algorithm can be used to compute
$\mc{P}_{\mrm{max}}(\{\mc{D}_i\},\sigma)$.

\begin{alg}
  \label{alg:stack-based}
  \textbf{Input:} The automata $\mc{D}_1, \dots, \mc{D}_n$ and the
  length-$N$ string $\sigma$.
  \begin{tabbing}
    \quad \= \quad \= \quad \= \quad \= \quad \= \quad \= \quad \+ \kill
    \textbf{Let} $\mc{A} := \Det(\mc{D}_1 \sqcup \cdots \sqcup
    \mc{D}_n)$. \\
    \textbf{Let} $s_0$ be $\mc{A}$'s unique start state. \\
    \textbf{Let} $\mathbf{S}$ and $\mathbf{M}$ be empty stacks. \\
    \textbf{For} $j = 1 \dots N$ \textbf{do} \+ \\
    Push $(s_0,j)$ onto $\mathbf{S}$. \\
    \textbf{For} each $(s,i) \in \mathbf{S}$ \textbf{do} \+ \\
    \textbf{If} there is a transition $\trans{s}{\sigma_j}{s'} \in
    T(\mc{A})$ \+ \\
    \textbf{then} replace $(s,i)$ with $(s',i)$ in $\mathbf{S}$. \\
    \textbf{Otherwise}, remove $(s,i)$ from $\mathbf{S}$. \+ \\
    \textbf{If}, in addition, $(s,i)$ was at the bottom
    of $\mathbf{S}$ \\
    \> \textbf{then} push the pair $(i,j-1)$ onto $\mathbf{M}$. \- \-
    \- \- \\
    \textbf{Let} $(s_f,i_f)$ be the pair at the bottom of
    $\mathbf{S}$. \\
    Push $(i_f,N)$ onto $\mathbf{M}$. \- \\
    \textbf{Output:} $\mathbf{M}$.
  \end{tabbing}
\end{alg}

The following proposition is easily verified, and we state it without
proof.
\begin{prop}
  \label{stack-local-prop}
  If $\sigma$ is a \emph{finite string} and if $\mathbf{M}_\sigma$ is
  the output of the above algorithm when applied to $\sigma$, then
  $\mc{P}_\mrm{max}(\{\mc{D}_i\},\sigma) = \{ \sigma_{a,b} : (a,b) \in
  \mathbf{M}_\sigma\}$.
\end{prop}

We summarize Prop.~\ref{stack-local-prop} by saying that
Algorithm~\ref{alg:stack-based} solves the \emph{local} filtering
problem in the sense that it can compute
$\mc{P}_\mrm{max}(\{\mc{D}_i\}, w)$ over a finite, contractible
window~$w$.  (By \emph{contractible} we mean that periodic boundary
conditions along the boundary of $w$ are ignored.)

The \emph{global} filtering problem, which takes into account periodic
boundary conditions, is considerably more subtle.  A somewhat pedantic
example is filtering the bi-infinite string $0^\Zr$ consisting
entirely of $0$s with the language $\sub[(0^m 1)^*]$.  (Recall that
$\sub(\mc{L})$ is our notation for the collection of substrings of
strings belonging to $\mc{L}$.)  The local approach applied to a
finite length-$N$ window $0^N$, where $N<m$, will return $0^N$ itself
as its single maximal substring; i.e., $\mc{P}_\mrm{max}(\{\sub[(0^m
1)^*]\},0^N) = \{ 0^N \}$.  In contrast, the global filter of $0^\Zr$
will consist of heavily overlapping length-$m$ substrings beginning
and ending at every position within $0^\Zr$:
\begin{align*}
  \mc{P}_\mrm{max}(\{\sub[(0^m 1)^*]\},0^\Zr) = \{ 0^\Zr_{a+1}
  0^\Zr_{a+2} \cdots 0^\Zr_{a+m} : a \in \Zr\}.
\end{align*}

Fortunately, by examining sufficiently large finite windows,
Algorithm~\ref{alg:stack-based} can also be used to solve this more
subtle global filtering problem in the case of a bi-infinite string
that is periodic.  The following Lemma captures the essential
observation.

\begin{lem}
  \label{lem:pumping}
  Suppose $\sigma$ is a \emph{period-$N$ bi-infinite string}.  Then
  every maximal substring $\sigma_{a,b} \in
  \mc{P}_\mrm{max}(\{\mc{D}_i\},\sigma)$ must have length $\le m \cdot
  N$, where $m := \max \{ |S(\mc{D}_i)| \}_i$, or else
  $\mc{P}_\mrm{max}(\{\mc{D}_i\},\sigma)$ must consist of
  $\sigma_{-\infty,\infty}=\sigma$, alone.
\end{lem}
\begin{proof}
  Our argument is a variation on the proof of the classical Pumping
  Lemma from automata theory.  Suppose that $\sigma_{a,b} \in
  \mc{P}_\mrm{max}(\{\mc{D}_i\},\sigma)$, $a$ and $b$ are finite,
  and $b-a+1 > m \cdot N$.  Then one of the domains, say
  $\mc{D}_i$, accepts $\sigma_{a,b}$.  By definition, this means there
  is a sequence of transitions in $T(\mc{D}_i)$ of the form
  $\trans{s_a}{\sigma_a}{s_{a+1}},
  \trans{s_{a+1}}{\sigma_{a+1}}{s_{a+2}}, \dots,
  \trans{s_b}{\sigma_b}{s_{b+1}}$.  Consider the sequence of pairs:
  \begin{align*}
    \{(s_i,i \bmod N)\}_{i=a}^b \subset S(\mc{D}_i) \times \Zr_N.
  \end{align*}
  Since: 
  \begin{align*}
    b-a+1 > m \cdot N \ge |S(\mc{D}_i) \times \Zr_N|,
  \end{align*}
  the Pigeonhole Principle implies that this sequence must
  repeat---say $(s_l, l \bmod N) = (s_{l'}, l' \bmod N)$ for integers
  $l<l'$.  But then $\mc{D}_i$ must also accept any string of the
  form:
  \begin{align*}
    \sigma_a \sigma_{a+1} \cdots \sigma_l (\sigma_{l+1} \cdots
    \sigma_{l'})^* \sigma_{l'+1} \cdots \sigma_b.
  \end{align*}
  Since $l \bmod N = l' \bmod N$, such strings correspond to
  arbitrarily long substrings of the original bi-infinite string
  $\sigma$.  As a result, $\sigma_{a,b}$ cannot be maximal.  This
  contradiction implies that either (i)~$a$ and $b$ are not both
  finite or (ii)~$b-a+1 \le m \cdot N$.  A straightforward
  generalization of our argument in fact shows that either
  (i)~\emph{both $a$ and $b$ are infinite} or (ii)~$b-a+1 \le m \cdot
  N$.
\end{proof}

A consequence of Lemma~\ref{lem:pumping} is that we can solve the
global filtering problem by applying Algorithm~\ref{alg:stack-based}
to a window of length $m N+1$.

\begin{prop}
  \label{stack-global-prop}
  Suppose $\sigma$ is a \emph{period-$N$ bi-infinite string} and that
  $\mathbf{M}_{\sigma'}$ is the output of
  Algorithm~\ref{alg:stack-based} when applied to the finite string
  $\sigma' := \sigma_1 \sigma_2 \cdots \sigma_{m N+1}$, where $m :=
  \max \{ |S(\mc{D}_i)| \}_i$.  Then:
  \begin{align*}
    \mc{P}_\mrm{max}(\{\mc{D}_i\},\sigma)
    = \{ \sigma_{a+qN,b+qN} : (a,b) \in
    \mathbf{M}_{\sigma'}, q \in \Zr \} ~,
  \end{align*}
  unless $\mathbf{M}_{\sigma'}$ consists of $(1,mN+1)$ alone, in which
  case $\mc{P}_\mrm{max}(\{\mc{D}_i\},\sigma) =
  \{\sigma_{-\infty,\infty}=\sigma\}$.
\end{prop}

The major drawback of Algorithm~\ref{alg:stack-based}, however, is its
worst-case compute time.

\begin{prop}
  \label{prop:stack-runtime}
  The worst-case performance of the stack-based filtering algorithm
  (Algorithm~\ref{alg:stack-based}) has order $\mc{O}(N^2)$, where $N$
  is the length of the input string~$\sigma$.
\end{prop}
\begin{proof}
  For each $j=1 \dots N$, the algorithm pushes a new pair $(s_0,j)$
  onto the stack $\mathbf{S}$ and then advances each pair on
  $\mathbf{S}$.  In the case that $\mc{A}$ accepts the entire string
  $\sigma$, the algorithm will never remove any pairs from
  $\mathbf{S}$ and will thus advance a total of $\sum_{j=1}^N j =
  \tfrac{1}{2}N(N+1)$ pairs.  The proposition follows since it is
  possible to advance each pair in constant time.
\end{proof}

\section{Method 2---Filtering with a Transducer}

The second method---and our primary focus---is to algorithmically
construct a finite transducer that approximates the stack-based
Algorithm~\ref{alg:stack-based} by printing sequences of labels
$i$ over segments of $\sigma$ recognized by the automaton~$\mc{D}_i$.
When, at the end of such a segment, the transducer encounters a letter
forbidden by the prevailing automaton~$\mc{D}_i$, it prints special
symbols until it resynchronizes to a new automaton $\mc{D}_j$. The
special symbols consist of labels for the kinds of domain-to-domain
transition and $\lambda$, which indicates that classification is
ambiguous.

In this way, the transducer approximates the stack-based algorithm by
jumping from one maximal substring to the next, printing a few special
symbols in between.  Because it does not jump to a new maximal
substring until the preceding one ends, however, the transducer can
miss the true beginning of any maximal substring that overlaps with
the preceding one.  But if no more than two maximal substrings overlap
at any given point of $\sigma$, then it is possible to combine the
output of two transducers, one reading left-to-right and the other
reading right-to-left, to obtain the same output as the stack-based
algorithm.

These shortcomings are minor, and in exchange the transducer gains
several significant advantages over the stack-based algorithm it
approximates: It requires only a finite amount of memory, runs in
linear time, and gives immediate output for each letter read.

Although finite transducers are generally considered less sophisticated
than stack-based algorithms in the sense of computational complexity, the
construction of this transducer is considerably more intricate than the
preceding stack-based algorithm and is, in fact, our principal aim in
the following.

Our approach will be to construct a transducer $\Filter(\{\mc{D}_i\})$
by `filling in' the forbidden transitions of the automaton $\mc{A} :=
\Det(\mc{D}_1 \sqcup \cdots \sqcup \mc{D}_n)$.  We will thus tie our
hands behind our backs at the outset by permitting the transducer to
remember \emph{only} as much about past input as does the automaton
$\mc{A}$ while recognizing domain strings.

Unfortunately, $\mc{A}$'s states will generally preserve too little
information to facilitate optimal resynchronization.  It is possible,
however, to begin with elaborately constructed, equivalent,
non-minimal domains $\mc{D}'_i$ that yield an automaton $\mc{A}'
:= \Det(\mc{D}'_1 \sqcup \dots \sqcup \mc{D}'_n)$ whose states do
preserve just enough information to facilitate optimal
resynchronization.  The transducer obtained by `filling in' the
forbidden transitions of this automaton $\mc{A}'$ represents the best
possible (transducer) approximation of the stack-based algorithm.  We
present a preprocessing algorithm which produces these equivalent,
non-minimal domains $\{\mc{D}'_i\} = \Optimize(\{\mc{D}_i\})$ at the
end of our
discussion of Method-2 filtering.

The idea underlying our construction is the following.  Suppose that
while reading the string $\sigma$ we are recognizing an increasingly
long string accepted by $\mc{D}_i$ when we encounter a forbidden
letter $a$.  In accepting $\sigma$ up to this point, the automaton
$\mc{A}$ will have reached a certain state $s \in S(\mc{A})$ that has
no outgoing transition corresponding to the letter $a$.  Our goal is
to create such a transition by examining the collection of all
possible strings that could have placed us in the state $s$ and to
resynchronize to the state of $\mc{A}$ that is most compatible with
the potentially foreign strings obtained by appending to these strings
the forbidden letter~$a$.

In this situation there will be two natural desires. On the one hand,
we wish to unambiguously resynchronize to as \emph{specific} a domain
state as possible; but, on the other, we wish to rely on as little of
the imagined past as possible. (We use the term \emph{imagined} because
our transducer remembers only the state $s \in S(\mc{A})$ we have
reached---not the particular string that placed us there.)  To reflect
these desires, we introduce a partial ordering on the collection of
potential resynchronization states $\{S_{i,l}\}$, where $i$ measures
the specificity of resynchronization and $l$ the length of imagined
past.

We now implement this intuition in full detail.  Our exposition relies
heavily on ideas from automata theory. (We now urge reading
App.~\ref{automata-review} in its entirety.)

\begin{figure}[here]
\begin{minipage}[t]{2.3in}
\includegraphics[width=2.3in]{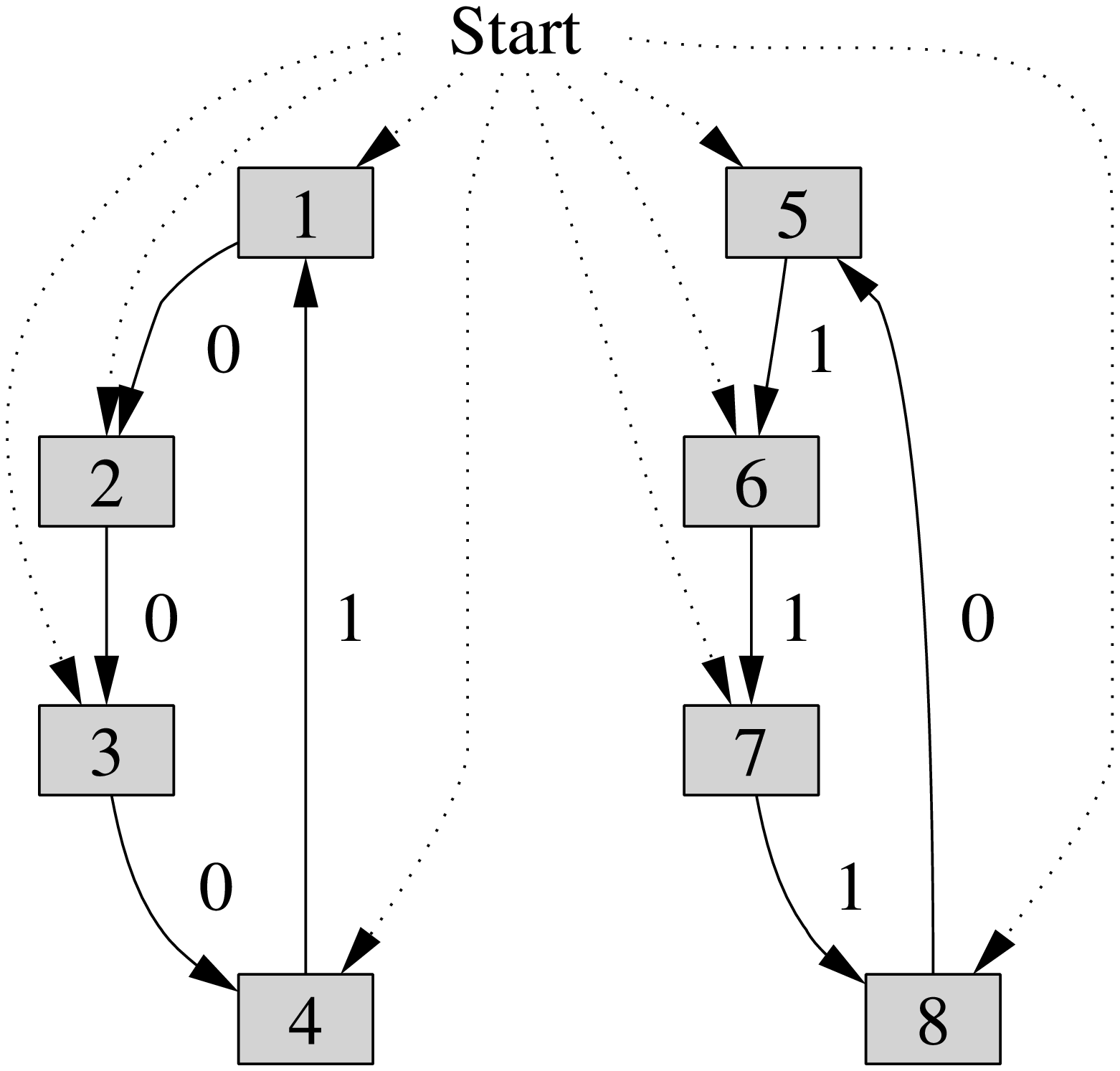}
\end{minipage}
\begin{minipage}[t]{2.4in}
\includegraphics[width=2.4in]{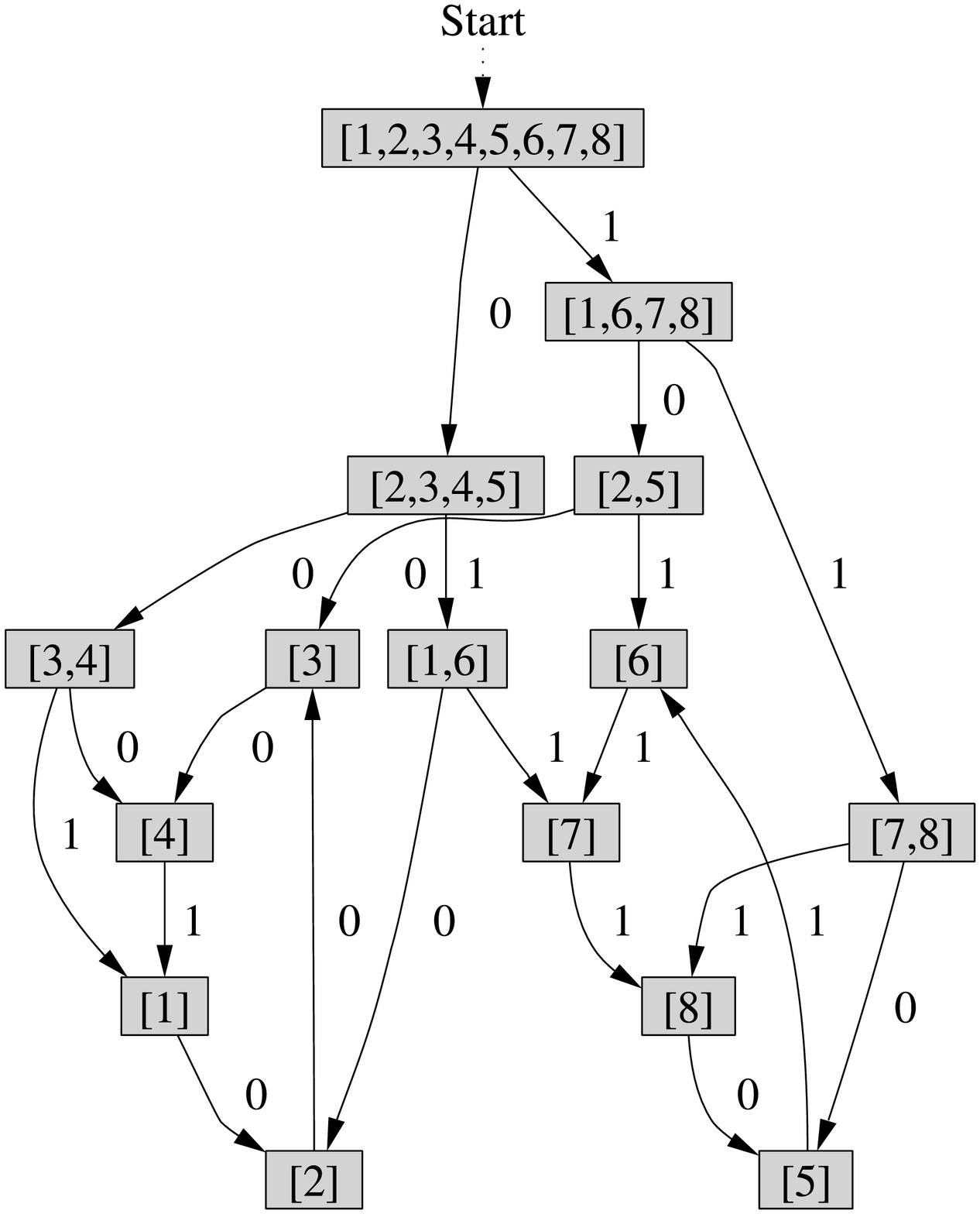}
\end{minipage}
\caption{\label{domains} The domains $\mc{D}_1$ and $\mc{D}_2$ (top) 
  and the automaton $\mc{A} = \Det(\mc{D}_1 \sqcup \mc{D}_2)$
  (bottom).  Start states are indicated by dotted arrows from the
  word ``Start'', and final states are darkened.  Notice that the
  states of $\mc{A}$ correspond to collections of states of $\mc{D}_1$
  and $\mc{D}_2$ and that the former are canonically injected into
  the latter, here by the map $n \mapsto [n]$.
  }
\end{figure}

As above, let $\mc{A} := \Det(\mc{D}_1 \sqcup \dots \sqcup \mc{D}_n)$
and let $S(\mc{A}) \stackrel{\psi_\mc{A}}{\hookrightarrow} S(\mc{D}_1
\sqcup \dots \sqcup \mc{D}_n)$ be the canonical injection provided by
Lemma~\vref{subset-construction} in App.~\ref{automata-review}.
Assume that there is a canonical injection $S(\mc{D}_1) \sqcup \dots
\sqcup S(\mc{D}_n) \hookrightarrow S(\mc{A})$ and that we can
therefore regard the sets $S(\mc{D}_i)$ as subsets of $S(\mc{A})$.  An
example of this situation is depicted in \Figure~\vref{domains}.  A
sufficient condition for the existence of such an injection is that
each $\mc{D}_i$ is minimal and that $\Lang(\mc{D}_i) \not\subset
\Lang(\mc{D}_{l})$ for $i \ne l$.  Minimality is far from required,
however, and the assumption is valid for a much larger class of
domains.  (Put informally, it suffices if we can associate to each
state $s \in S(\mc{D}_1 \sqcup \dots \sqcup \mc{D}_n)$ a string that
corresponds to a unique path through $\mc{D}_1 \sqcup \cdots \sqcup
\mc{D}_n$---one that leads to~$s$.)

Let $\mc{T}$ be a transducer with the same states, start state, and
final states as $\mc{A}$, but with the transitions:
\begin{align*}
  T(\mc{T}) :=& \{ \trans{s}{a|f(s')}{s'} : \trans{s}{a}{s'} \in
  T(\mc{A}) \} ~,
\end{align*}
where:
\begin{align*}
  f(s') = 
  \begin{cases}
    i & \text{if $\phi_\mc{A}(s') \subset S(\mc{D}_i)$} ~,\\
    \lambda & \text{otherwise} ~,
  \end{cases}
\end{align*}
and where $\lambda$ is a new symbol in the output alphabet $\Sigma'$
indicating that domain labeling was not possible, for example,
because the partial string read so far belongs to more than one or
none of the automata $\{\mc{D}_i\}$. To recapitulate, the transducer's
output alphabet $\Sigma'$ consists of three kinds of symbol: domain
labels $\{1 \ldots n\}$, domain-domain transition types
$\{1, 2, \ldots, p\}$, and ambiguity $\lambda$.

The transducer $\mc{T}$'s input, $\Input(\mc{T})$, recognizes
precisely those strings recognized by the given domains.  Our goal is
to extend $\mc{T}$ by introducing transitions of the form:
\begin{align*}
  \{ \trans{s}{a|h(s,a)}{g(s,a)} \ : \ & s \in S(\mc{T}) = S(\mc{A}),
  a \in \Sigma, \\ & \text{and there are no transitions} \\ & \text{of
    the form $\trans{s}{a}{\cdot} \in T(\mc{A})$} \} ~,
\end{align*}
where the functions $g(s,a)$ and $h(s,a)$ are defined in the following
paragraphs.  The transducer $\Filter(\{\mc{D}_i\})$ obtained by adding
these transitions to $\mc{T}$ will then have the desired property that
its input $\Input(\Filter(\{\mc{D}_i\}))$ will accept all
strings~\footnote{Accepting $\Sigma^*$ is rather generous: the filter
  will take any string and label its symbols according to the
  hypothesized patterns $\{\mc{D}_i\}$. This is not required for
  cellular automata, since their configurations often contract to
  strict subsets of $\Sigma^*$ over time.}.

\begin{figure}[here]
\begin{minipage}[t]{3.5in}
\includegraphics[width=3.5in]{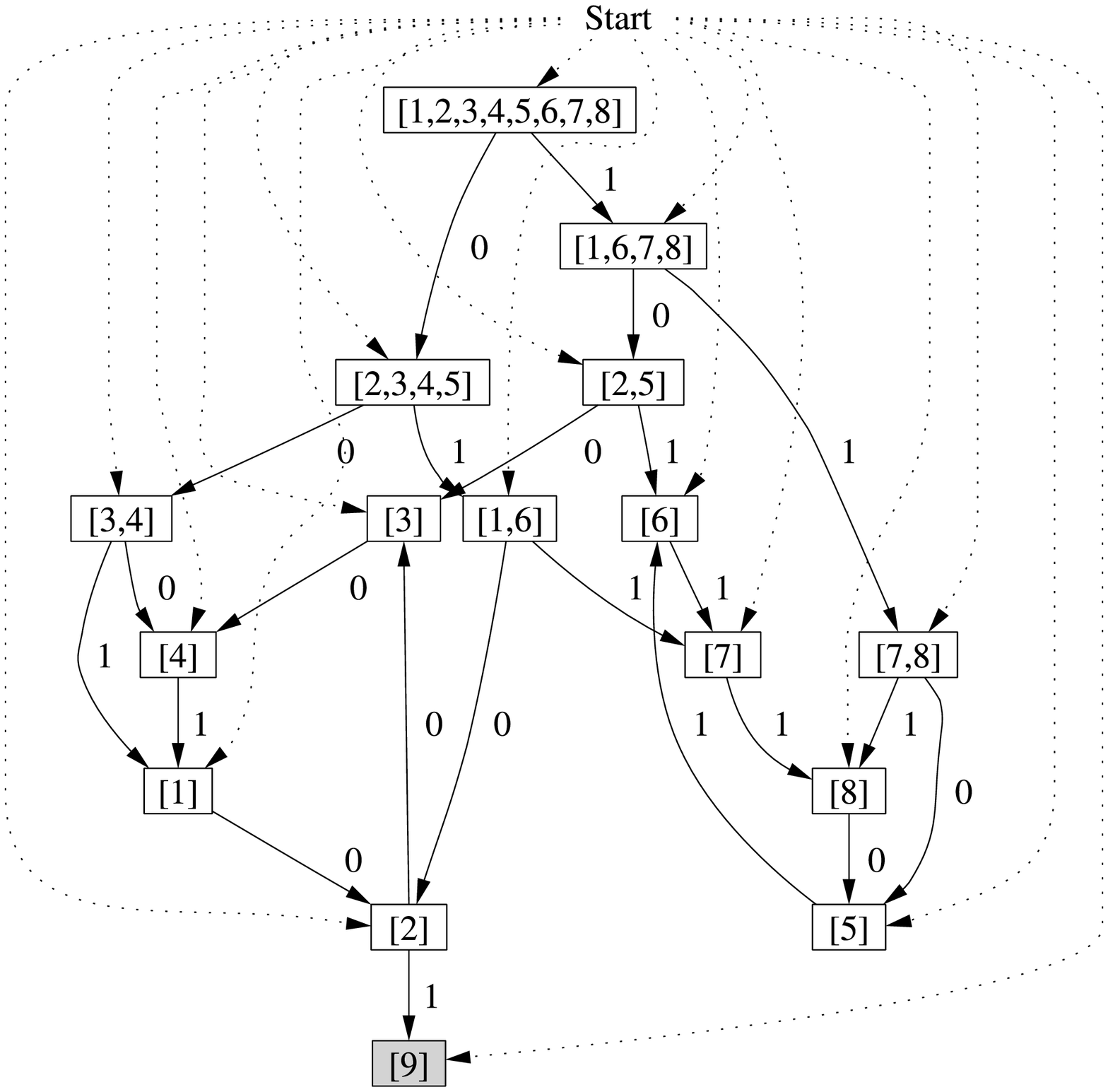}
\end{minipage}
\begin{minipage}[t]{2.4in}
\includegraphics[width=2.4in]{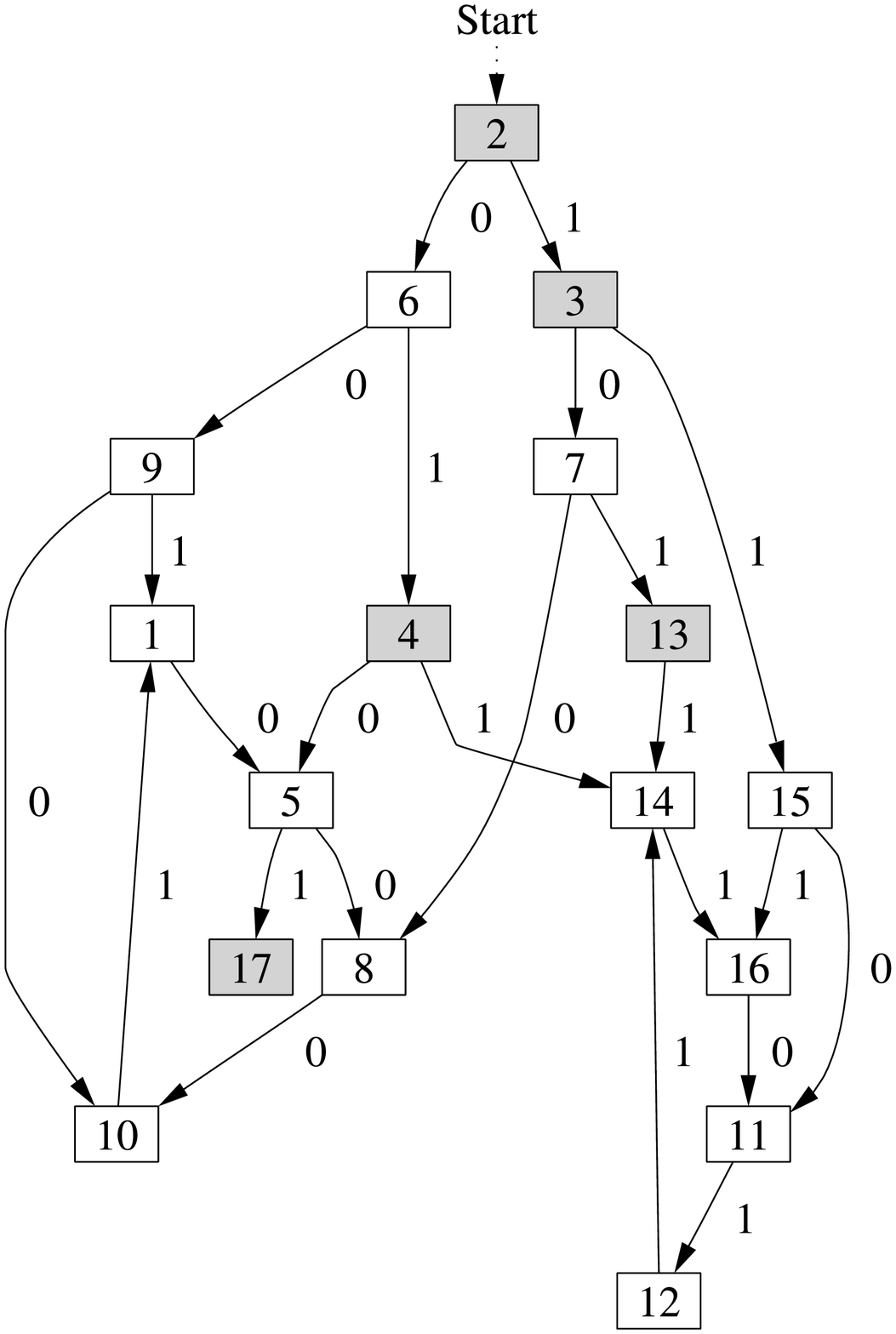}
\end{minipage}
\caption{\label{deadend}The semi-deterministic automaton
  $\mc{A}^{[2],1}$ (top) obtained by adding a state $f=[9]$ and its
  deterministic version $\Det(\mc{A}^{[2],1})$ (bottom) with states
  relabeled with the integers $1\dots 17$ in order to simplify later
  diagrams.
  }
\end{figure}

Let $W_l$ denote the collection of strings corresponding to length-$l$
paths through $\mc{A}$ beginning in any of its states, but
\emph{ending} in state~$s$, and let $W'_{l+1}$ denote the collection
of strings obtained by appending the letter $a$ to the strings of
$W_l$.  The strings $\bigcup_{l \ge 0} W'_l$ are accepted by the
finite automaton $\mc{A}^{s,a}$ obtained by adding a new state $f$ and
a transition $\trans{s}{a}{f}$ to $\mc{A}$, and by setting
$\mathrm{Start}(\mc{A}^{s,a}) := S(\mc{A}^{s,a})$ and
$\mathrm{Final}(\mc{A}^{s,a}) := \{f\}$.  An example is shown in
\Figure~\vref{deadend}, where the four-state domain has a transition
added from state~[2] on symbol~$1$, which was originally forbidden.

In order to choose the resynchronization state $g(s,a)$ for the
forbidden transition $(s,a)$, we examine the strings of $\bigcup_{l
  \ge 0} W'_l$ that also belong to one or more of the domains
$\{\mc{D}_i\}$.  We do this by constructing the automaton
$\Det(\mc{A}^{s,a}) \cap \mc{A}$, which we call the
\emph{resynchronization automaton}.  By
Lemma~\vref{intersection-construction}, there is a canonical, although
not necessarily injective, association:
\begin{equation*}
  \phi : S(\Det(\mc{A}^{s,a}) \cap \mc{A}) \rightarrow S(\mc{A})
\end{equation*}
given by the composition:
\begin{equation*}
S(\Det(\mc{A}^{s,a}) \cap \mc{A}) \ \ \hookrightarrow \ \
S(\Det(\mc{A}^{s,a})) \times S(\mc{A}) \ \ \rightarrow \ \ S(\mc{A}) ~,
\end{equation*}
where the right-most map is the second-factor projection, $(s,s')
\mapsto s'$.

The resynchronization automaton $\Det(\mc{A}^{s,a}) \cap \mc{A}$ may
reveal several possible resynchronization states.  To help distinguish
among them, we put them into sets $\{S_{i,l}\}$ where $i$
measures the \emph{specificity} of resynchronization and $l$ the length
of imagined past.  More precisely, let $S_{i,l}$ denote those states $s
\in S(\mc{A})$ to which $\phi$ associates at least one state $s' \in
\mathrm{Final}(\Det(\mc{A}^{s,a}) \cap \mc{A})$ (i.e.  $s=\phi(s')$)
satisfying the following two conditions: (1)~$s$ corresponds, under
Lemma~\vref{subset-construction}, to precisely $i$ states of $\mc{D}_1
\sqcup \cdots \sqcup \mc{D}_n$ and (2)~there is a length-$l$ path
from the unique start state of $\Det(\mc{A}^{s,a}) \cap \mc{A}$ to
$s'$.

\begin{figure}[here]
\begin{minipage}[t]{2.5in}
\includegraphics[width=2.5in]{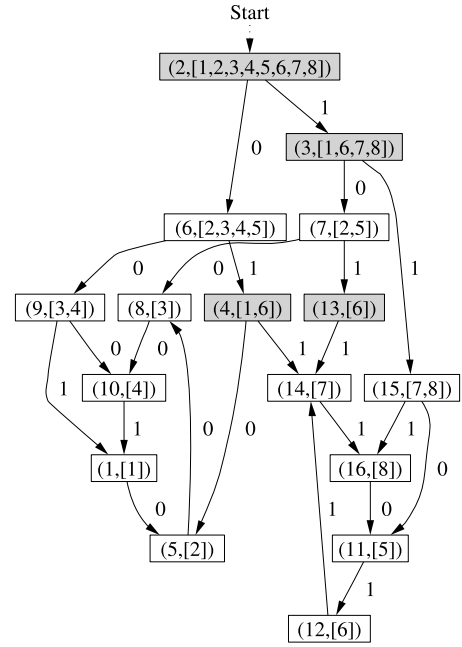}
\end{minipage}
\begin{minipage}[t]{2.5in}
\includegraphics[width=2.5in]{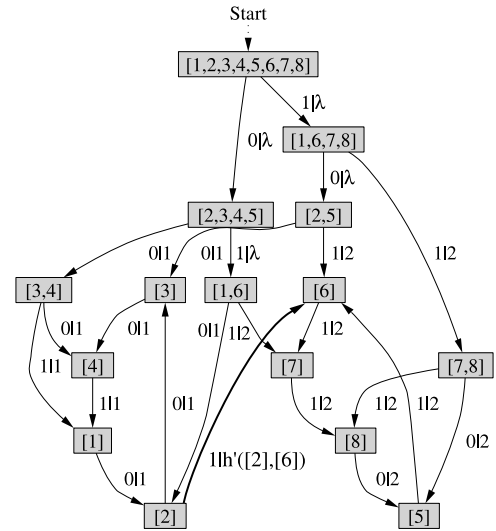}
\end{minipage}
\caption{\label{intersect}The resynchronization automaton 
  $\Det(\mc{A}^{[2],1}) \cap \mc{A}$ (top).  Here $S_{1,3}$ consists
  of the state $(13,[6])$ alone, and all other $S_{1,\bullet}$ are
  empty. So we choose $s' = [6]$ and add a transition
  $\trans{[2]}{1|h'([2],[6])}{[6]}$ to~$\mc{T}$ (bottom).}
\end{figure}

Give the sets $\{S_{i,l}\}$ the dictionary ordering; that is, let
$S_{i,l} < S_{i',l'}$ if $i<i'$ or if $i=i' \wedge l<l'$.  The set
$S_{|S(\mc{A})|,0}$ consists of the unique start state of
$\Det(\mc{A}^{s,a}) \cap \mc{A}$.  Thus, by the well ordering
principle, there must be a unique, least set among the sets
$\{S_{i,l}\}$ that consist of a single state, say $\{s'\}$.  Let
$g(s,a) := s'$, and let $h(s,a):=h'(s,s')=h'(s,g(s,a))$, where $h'$ is
any injection $S(\mc{T}) \times S(\mc{T}) \hookrightarrow \Sigma'$
(chosen independent of $s$ and $a$).  An example of this construction
is shown in \Figure~\vref{intersect}.

The transducer is completed by repeating the above steps for all
forbidden transitions.

\subsection*{Computability of the transducer $\Filter(\{\mc{D}_i\})$}

Although the transducer $\Filter(\{\mc{D}_i\})$ is well defined, it is
perhaps not immediately clear that it is computable.  After all, we
appealed to the well ordering principle to obtain a least singleton
set $\{s'\}$ among the sets $\{S_{i,l}\}$.  In fact, infinitely many
sets $S_{i,l}$ precede the stated upper bound
$S_{|S(\mc{A})|,0}$---for instance, all of the sets $S_{1,\Nn}$ do,
provided $|S(\mc{A})|>1$.

The construction is nevertheless computable, because for each $i$ the
sequence of sets $S_{i,\Nn}$ must eventually repeat.  In fact, we can
compute this sequence of sets exactly by automata-theoretic means.

\begin{prop}
  The transducer $\Filter(\{\mc{D}_i\})$ is computable.
\end{prop}
\begin{proof}
  Let $\mc{Z}[\mc{C}]$ denote the automaton obtained by relabeling all
  of the automaton $\mc{C}$'s transitions with $0$s.  This automaton
  will almost certainly be nondeterministic.  The equivalent
  deterministic automaton $\Det(\mc{Z}[\mc{C}])$ is useful, because
  the state it reaches when accepting the string $0^l$ corresponds
  precisely, under Lemma~\ref{subset-construction}, to the collection
  of states that can be reached by length-$l$ paths through $\mc{C}$.
  
  Moreover, since $\Det(\mc{Z}[\mc{C}])$ is defined over a single
  letter, yet deterministic and finite, it must have a special
  graphical structure: its single start state $s_0$ must lead to a
  finite loop after a finite chain of non-recurrent states.
  (Actually, if $\mc{C}$ has no loops whatsoever, there will not even
  be a loop.)  Thus, its states have a linear ordering: $s_0
  \stackrel{0}{\rightarrow} s_1 \stackrel{0}{\rightarrow} \cdots
  \stackrel{0}{\rightarrow} s_m \stackrel{0}{\rightarrow} s_{m+1}
  \stackrel{0}{\rightarrow} \cdots \stackrel{0}{\rightarrow} s_{m+m'}
  \stackrel{0}{\rightarrow} s_m$.  An example is illustrated in
  \Figure~\vref{squash}, where $m=4$ and $m'=0$.
  
  By Lemma~\vref{subset-construction} the states $\{s_k\}$ correspond
  to collections of states of $\mc{C}$ under an injection:
  \begin{align*}
    \psi_{\mc{Z}[\mc{C}]} : S(\Det(\mc{Z}[\mc{C}]))
    \hookrightarrow & \{ S \subset
    S(\mc{Z}[\mc{C}]) \}\\ =&  \{ S \subset
    S(\mc{C}) \}.
  \end{align*}
  
  Let $\mc{C} := \Det(\mc{A}^{s,a}) \cap \mc{A}$ in the preceding
  discussion.  As before, by Lemma~\vref{intersection-construction},
  there is a function:
  \begin{align*}
    \phi : S(\Det(\mc{A}^{s,a}) \cap \mc{A}) \rightarrow S(\mc{A}) ~.
  \end{align*}
  Let $S_{*,l} \subset S(\mc{A})$ denote those states defined by
  the formula:
  \begin{align*}
    S_{*,l} := \phi[\psi_{\mc{Z}[\mc{C}]}(s_l) \cap
    \mathrm{Final}(\Det(\mc{A}^{s,a}) \cap \mc{A})] ~.
  \end{align*}

  \begin{figure}[here]
    \begin{minipage}[t]{2.6in}
      \includegraphics[width=2.6in]{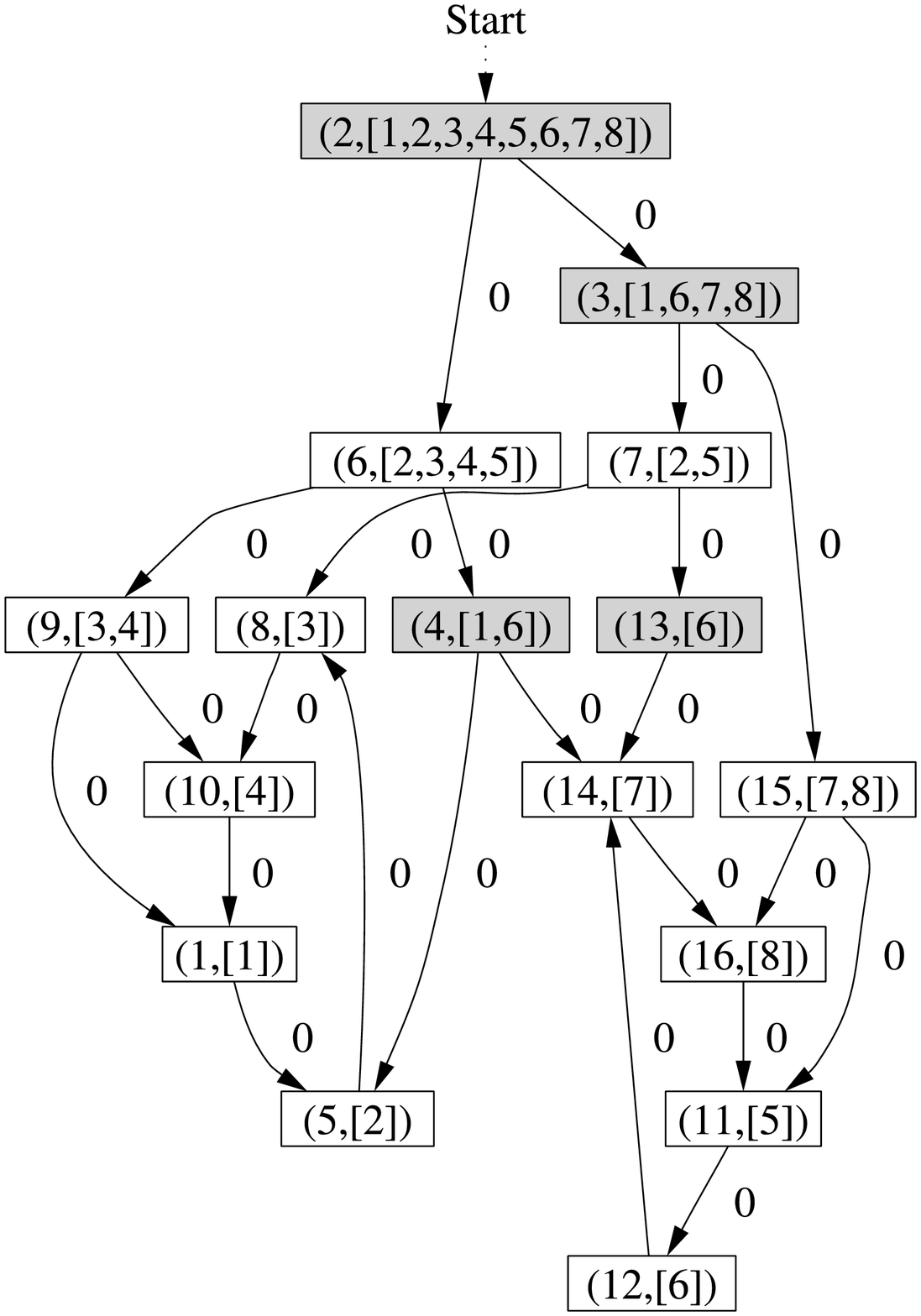}
    \end{minipage}
    \begin{minipage}[t]{3.1in}
      \includegraphics[width=3.1in]{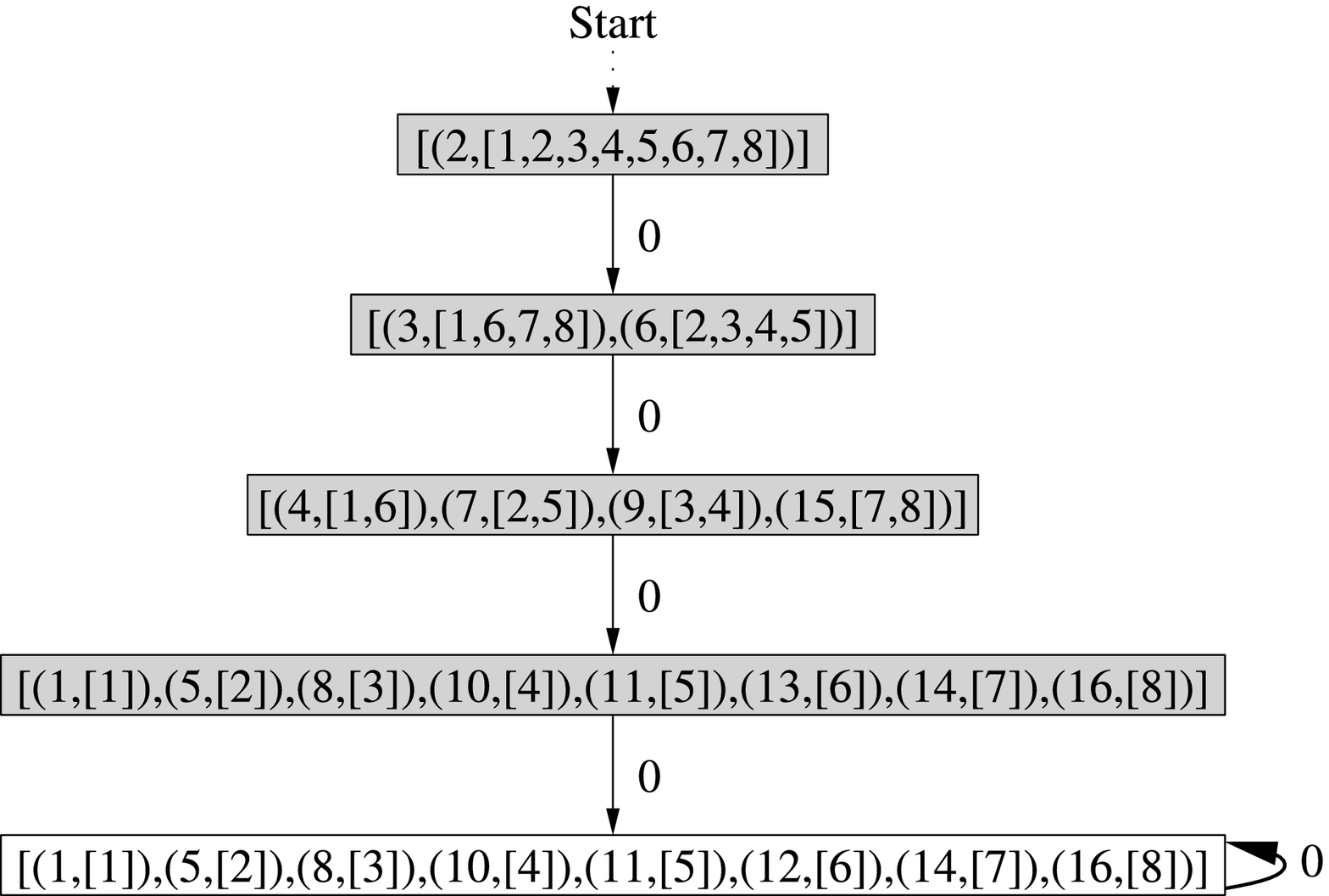}
    \end{minipage}
    \caption{\label{squash} The automaton 
      $\mc{Z}[\Det(\mc{D}_1^{[2],1}) \cap \mc{A}]$ (top) and its
      deterministic version $\Det(\mc{Z}[\Det(\mc{D}_1^{[2],1}) \cap
      \mc{A}])$ (bottom).}
  \end{figure}
  
  Finally, let $S_{i,*}$ denote those states of $\mc{A}$ that
  correspond to precisely $i$~states of $\mc{D}_1 \sqcup \cdots \sqcup
  \mc{D}_n$; that is, let:
  \begin{align*}
    S_{i,*} := \{ s \in S(\mc{A}) : |\phi_{\mc{A}}(s)| = i \} ~,
  \end{align*}
  where $\phi_{\mc{A}} : S(\mc{A}) \hookrightarrow \{ S \subset
  S(\mc{D}_1 \sqcup \cdots \sqcup \mc{D}_n) \}$ is the injection
  provided by Lemma~\vref{subset-construction}.
  
  The sets $S_{i,l}$ can then be computed as the intersections
  $S_{i,*} \cap S_{*,l}$, and we need only examine these for $1 \le i
  \le |S(\mc{D}_1)|+\dots+|S(\mc{D}_n)|$ and $0 \le l \le m+m'$ to
  discover the least one under the dictionary ordering that is a
  singleton $\{s'\}$.
\end{proof}

We summarize the entire algorithm.

{
\newenvironment{entry}
  {\begin{list}{--}{
      \setlength{\topsep}{0pt}
      \setlength{\itemsep}{0pt}
      \setlength{\parsep}{0pt}
      \setlength{\labelwidth}{5pt}
      \setlength{\itemindent}{0pt}}}{\end{list}}

\begin{alg}
  \label{alg:transducer-building}
  \textbf{Input:} The regular domains $\mc{D}_1, \dots, \mc{D}_n$.
  \begin{entry}
  \item Let $\mc{A} := \Det(\mc{D}_1 \sqcup \cdots \sqcup \mc{D}_n)$.
  \item Choose any injection $h' : S(\mc{A}) \times S(\mc{A})
    \hookrightarrow \Sigma'$.
  \item Make $\mc{A}$ into a transducer $\mc{T}$ by adding the symbol
    $i$ as output to any transition ending in a state corresponding to
    states of only one domain $\mc{D}_i$ and by adding $\lambda$s as
    output symbols to all other transitions.
  \item \textbf{For each} forbidden transition $(s,a) \in S(\mc{A})
    \times \Sigma(\mc{A})$, add a transition to $\mc{T}$ through the
    following procedure \textbf{do}
    \begin{entry}
    \item Construct the automaton $\mc{A}^{s,a}$ by adding to $\mc{A}$
      the transition $\trans{s}{a}{f}$, where $f$ is a new state, and
      by letting $f$ be its only final state.
    \item Construct the automaton $\Det(\mc{Z}[\Det(\mc{A}^{s,a}) \cap
      \mc{A}])$, where $\mc{Z}[\mc{C}]$ is the automaton obtained by
      relabeling all of $\mc{C}$'s transitions with $0$s.  Its states
      will have a natural linear ordering $s_0 \rightarrow s_1
      \rightarrow \cdots \rightarrow s_{m+m'}.$
    \item Let $S_{i,*}$ and $S_{*,l}$ be the subsets of $S(\mc{A})$
      defined by:
      \begin{align*}
        S_{*,l} &:= \phi[\psi_{\Det(\mc{Z}[\Det(\mc{A}^{s,a}) \cap
          \mc{A}])}(s_l) \\
        & \; \; \; \; \; \; \; \; \; \; \cap
        \mathrm{Final}(\Det(\mc{A}^{s,a}) \cap \mc{A})] \text{\ and} \\
        S_{i,*} &:= \{ s \in S(\mc{A}) : |\psi_{\mc{A}}(s)| = i \}.
      \end{align*}
    \item Find the singleton set $\{s'\}$ among the sets:
      \begin{align*}
        \{S_{i,*} \cap S_{*,l} : \; & 1 \le i \le |\mc{D}_1| + \cdots +
        |\mc{D}_n|, \\ & 0 \le l \le m+m' \}
      \end{align*}
      that occurs first under the dictionary ordering.
    \item Add the transition $\trans{s}{a|h'(s,s')}{s'}$ to~$\mc{T}$.
    \end{entry}
  \end{entry}
  \textbf{Output:} $\Filter(\{\mc{D}_i\}) := \mc{T}$.
\end{alg}
}

\subsubsection*{Algorithmic Complexity}

\begin{prop}
  The worst-case performance of the transducer-constructing algorithm
  (Algorithm~\ref{alg:transducer-building}) has order no greater than:
  \begin{align*}
    |\mc{A}| \cdot (|\Sigma|-1) \cdot \exp \circ \exp \left( 2 \cdot
      |\mc{A}| + 1 \right) ~,
  \end{align*}
  where $|\mc{A}|$ has order $\exp( |\mc{D}_1| + \cdots +
  |\mc{D}_n|)$.
\end{prop}
\begin{proof}
  The algorithm's most expensive step is the computation of
  $\Det(\mc{Z}[\Det(\mc{A}^{s,a}) \cap \mc{A}])$.  Unfortunately,
  because computing $\Det(\mc{G})$ has order $\mathrm{exp}(|\mc{G}|)$,
  and because computing $\mc{G} \cap \mc{H}$ has order $|\mc{G}| \cdot
  |\mc{H}|$, this computation has order $\exp \circ \exp \left( 2
    \cdot |\mc{A}| + 1 \right)$.
  
  Finally, recall that the algorithm computes
  $\Det(\mc{Z}[\Det(\mc{A}^{s,a}) \cap \mc{A}])$ for every forbidden
  transition $(s,a)$ of $\mc{A}$.  A rough upper bound for the number
  of such transitions is $|\mc{A}| \cdot (|\Sigma|-1)$.
  From these two upper bounds the proposition follows.
\end{proof}

Although this analysis may at first seem to objurgate the
transducer-constructing algorithm, the reader should realize that,
once computed, $\mc{T}$ can be very efficiently used to filter
arbitrarily long strings.  That is, unlike the stack-based algorithm,
its performance is linear in string length.  Thus, one pays during the
filter design phase for an efficient run-time algorithm---a trade-off
familiar, for example, in data compression.

\subsection*{Constructing optimal transducers from non-minimal
  domains, a preprocessing step to
  Algorithm~\ref{alg:transducer-building}}

\label{domain-preprocessing}

Recall that we constructed the transducer $\Filter(\{\mc{D}_i\})$ by
`filling in' the forbidden transitions of the automaton $\mc{A} :=
\Det(\mc{D}_1 \sqcup \dots \sqcup \mc{D}_n)$.  This proved somewhat
problematic, however, because $\mc{A}$'s states do not always preserve
enough information about past input to unambiguously resynchronize to
a unique, recurrent domain state.  In order to help discriminate among
the several possible resynchronization states, we introduced the
partially ordered sets $\{S_{i,l}\}$.  But even so, several attractive
resynchronization states often fell into the same set $S_{i,l}$.  So,
lacking any objective way to choose among them, we resigned ourselves
to a less attractive resynchronization state occurring in a later set
$S_{i',l'}$, simply because it appeared alone there, making our choice
unambiguous.  If only the states of the automaton $\mc{A}$ preserved
slightly more information about past input, then such compromises
could be avoided.

In this section we present an algorithm that splits the states of a
given collection $\{\mc{D}_i\}$ of domains to obtain an equivalent
collection $\{\mc{D}'_i\} = \Optimize(\{\mc{D}_i\})$ of domains that
preserve just enough information about past input to enable unambiguous
resynchronization in the transducer obtained by \emph{filling in} the
forbidden transitions of the automaton
$\mc{A}' := \Det(\mc{D}'_1 \sqcup \dots \sqcup \mc{D}'_n)$.

We will accomplish this by associating to each state of the original
domains $\mc{D}_i$ a collection of automata that partition past
input strings into equivalence classes corresponding to individual
resynchronization states.  We will then refine these partitions so
that $\mc{D}_i$'s transition structures can be lifted to them and
thus obtain the desired domains $\{\mc{D}'_i\} =
\Optimize(\{\mc{D}_i\})$.

This procedure, taken as a preprocessing step to
Algorithm~\ref{alg:transducer-building}, will thus produce the best
possible transducer for Method-2 multi-regular language filtering.

We now state our construction formally.  If $s' \in S(\mc{A})$, then
let $\mc{A}_{s'}$ denote the automaton that is identical to the
automaton $\mc{A}$ except that its only final state is $s'$.
Additionally, if $(s,a)$ is a forbidden transition of the automaton
$\mc{D}_1 \sqcup \dots \sqcup \mc{D}_n$, then let $\mc{B}(s,a,s')$
denote the automaton satisfying the formula:
\begin{align*}
  \mc{B}(s,a,s') \cdot a = \Det(\mc{A}^{s,a}) \cap \mc{A}_{s'} ~,
\end{align*}
where $\cdot$ denotes concatenation.  That is, let $\mc{B}(s,a,s')$
denote the automaton that is identical to the automaton
$\Det(\mc{A}^{s,a}) \cap \mc{A}_{s'}$ except that its final states are
given by $\{ s_f : \trans{s_f}{a}{s_f'} \in T(\diamond), s_f' \in
\mrm{Final}(\diamond)\}$, where $\diamond := \Det(\mc{A}^{s,a}) \cap
\mc{A}_{s'}$.  Note that in most cases $\Lang(\mc{B}(s,a,s'))$ will be
empty.

Next we associate to each state $s \in S(\mc{D}_1 \sqcup \dots \sqcup
\mc{D}_n)$ a collection $\Gamma(s)$ of automata.  If the state $s$ has
no forbidden transitions, let $\Gamma(s) := \{ \Sigma^* \}$.  If the
state $s$ has at least one forbidden transition, however, then let
$\Gamma(s)$ denote the collection of automata:
\begin{align*}
  \Gamma(s) := \mrm{Disjoin}(\{ & \Sigma^* \cdot \mc{B}(s,a,s') : \\ &
  \trans{s}{a}{\cdot} \not\in T(\mc{A}), s' \in S(\mc{A})\}) ~,
\end{align*}
where $\mrm{Disjoin}(\{\mc{C}_\gamma\})$ denotes the coarsest
partition of $\bigcup_\gamma \Lang(\mc{C}_\gamma)$ by automata
$\{\mc{E}_\epsilon\}$ that is compatible with the automata
$\{\mc{C}_\gamma\}$. That is, $\mrm{Disjoin}(\{\mc{C}_\gamma\})$
denotes the smallest collection $\{\mc{E}_\epsilon\}$ of automata
satisfying (i)~$\bigcup_\epsilon \Lang(\mc{E}_\epsilon) =
\bigcup_\gamma \Lang(C_\gamma)$ and (ii)~$\Lang(\mc{C}_\gamma) \cap
\Lang(\mc{E}_\epsilon)$ is either empty or equal to
$\Lang(\mc{E}_\epsilon)$ for all $\gamma$ and $\epsilon$.

It is possible to compute $\mrm{Disjoin}(\{\mc{C}_\gamma\})$ inductively
with the formula:
\begin{align*}
  \mrm{Disjoin}&(\{\mc{C}_1, \mc{C}_2, \dots, \mc{C}_m\}) = \\ & \{
  \mc{C}_1 \setminus (\mc{C}_2 \sqcup \cdots \sqcup \mc{C}_m)\} \;
  \cup \\ &\{ \mc{C}_1 \cap \mc{C}' : \mc{C}' \in
  \mrm{Disjoin}(\{\mc{C}_2, \dots, \mc{C}_m\}) \} \; \cup \\ & \{
  \mc{C}_1 \setminus \mc{C}' \, : \mc{C}' \in \mrm{Disjoin}(\{\mc{C}_2,
  \dots, \mc{C}_m\}) \}.
\end{align*}


Note that $\bigcup_{\mc{E} \in \Gamma(s)} \Lang(\mc{E}) = \Sigma^*$
for all states $s \in S(\mc{D}_1 \sqcup \dots \sqcup \mc{D}_n)$.  This
is because $\Lang(\mc{B}(s,a,s'))$ contains only the empty string if
$s' \in S(\mc{A})$ is the unique state reached on input $a$ from
$\mc{A}$'s starting state---that is, if $(s_0,a,s') \in T(\mc{A})$,
where $\{s_0\} = \mrm{Start}(\mc{A})$.

Our goal is to create for each original domain $\mc{D}_i$ an
equivalent domain $\mc{D}'_i$ by splitting each state $s \in
S(\mc{D}_i)$ into states of the form $(s,\mc{E})$, where $\mc{E} \in
\Gamma(s)$.  But to endow these split states with a transition
structure equivalent to $\mc{D}_i$'s, we typically must refine the
sets $\Gamma(s)$ further. We must construct a refinement $\Gamma'(s)$
of each $\Gamma(s)$ with the property that if $\trans{s}{a}{s'}$ is a
transition of $\mc{D}_i$, then to each $\mc{E} \in \Gamma'(s)$ there
corresponds a unique $\mc{E}' \in \Gamma'(s')$ with $\Lang(\mc{E}
\cdot a) \subset \Lang(\mc{E}')$.  Given such refinements
$\Gamma'(s)$, we can take the pairs $\{(s,\mc{E}) : s \in S(\mc{D}_i),
\mc{E} \in \Gamma'(s)\}$ as the states of $\mc{D}'_i$ and equip them
with transitions of the form $(s,\mc{E}) \stackrel{a}{\rightarrow}
(s',\mc{E}')$, and thus obtain an equivalent, but non-minimal,
domain~$\mc{D}'_i$.

The following algorithm can be used to compute the desired refinements
$\Gamma'(s)$.

{
\newenvironment{entry}
  {\begin{list}{--}{
      \setlength{\topsep}{0pt}
      \setlength{\itemsep}{0pt}
      \setlength{\parsep}{0pt}
      \setlength{\labelwidth}{5pt}
      \setlength{\itemindent}{0pt}}}{\end{list}}

\begin{alg}
  \label{alg:optimize-domains}
  \textbf{Input:} The domain $\mc{D}$ and the function~$\Gamma$ that
  assigns to each state $s \in S(\mc{D})$ a collection $\Gamma(s)$ of
  automata that partition $\Sigma^*$.
  \begin{entry}
  \item \textbf{For each} state $s \in S(\mc{D})$, \textbf{let}:
    \begin{align*}
      \Gamma'(s) &:= \mrm{Disjoin}\left( \bigcup
        \{ \Gamma'(s,a,s') : (s,a,s') \in T(\mc{D}) \} \right) ~,
    \end{align*}
    where:
    \begin{align*}  
      \Gamma'(s,a,s') &:= \{ \mc{E}_\alpha'' \}_\alpha \text{\ and} \\
      \{\mc{E}_\alpha'' \cdot a\}_\alpha &:= \{ (\mc{E} \cdot a) \cap
      \mc{E}' : \mc{E} \in \Gamma(s), \mc{E}' \in \Gamma(s') \}.
    \end{align*}
  \item If $\Gamma'(s) \ne \Gamma(s)$ for some state $s \in
    S(\mc{D})$, then \textbf{repeat} with $\Gamma'$ in place
    of $\Gamma$.  Otherwise: \\
  \end{entry}
  \textbf{Output:} $\Gamma'$.
\end{alg}
}

\begin{prop}
  \label{prop:optimization-terminates}
  Algorithm~\vref{alg:optimize-domains} eventually terminates,
  producing the coarsest possible refinements $\Gamma'(s)$ of
  $\Gamma(s)$ compatible with $\mc{D}$'s transition structure.
\end{prop}
\begin{proof}
  We construct fine, but finite, refinements that are compatible with
  $\mc{D}$'s transition structure, then use this result to conclude
  that Algorithm~\vref{alg:optimize-domains} must eventually terminate.
  Moreover, we also conclude that, when Algorithm~\ref{alg:optimize-domains}
  terminates, it produces the coarsest possible refinements that are
  compatible with $\mc{D}$'s transition structure.
  
  Let $\{\mc{E}_i\}$ denote the potentially large, but finite,
  collection of automata:
  \begin{align*}
    \{\mc{E}_i\}_{i=1}^N := \mrm{Disjoin}\left( \bigcup_{s \in
        S(\mc{D})} \! \! \! \Gamma(s) \right) ~,
  \end{align*}
  which partition $\Sigma^*$.
  
  We refine the partition $\{\mc{E}_i\}$ to make it compatible with
  $\mc{D}$'s transitions by examining the automaton $\mc{F} :=
  \Det(\mc{E}_1 \sqcup \dots \sqcup \mc{E}_N)$. Since the automata
  $\{\mc{E}_i\}$ cover $\Sigma^*$, the deterministic automaton
  $\mc{F}$ can have no forbidden transitions, and all its states must
  be final.  Moreover, because the automata $\{\mc{E}_i\}$ are
  disjoint, each of $\mc{F}$'s states must correspond (under the
  canonical injection $\psi_\mc{F}$ of
  Lemma~\vref{subset-construction}) to final states of precisely one
  automaton $\mc{E}_i$.  In this way, the automata $\{\mc{E}_i\}$
  correspond to a partition of the states of $\mc{F}$.
  
  Since each automaton $\mc{E}_i$ is equivalent to the automaton
  obtained by restricting $\mc{F}$'s final states to those states
  corresponding (under $\psi_\mc{F}$) to final states of $\mc{E}_i$,
  we can refine the partition $\{\mc{E}_i\}$ by refining this
  partition of $\mc{F}$'s states.
  
  Although a coarser refinement may suffice, we can always choose the
  partition consisting of single states.  That is, if $s \in
  S(\mc{F})$, let $\mc{F}_s$ denote the automaton that is identical to
  the automaton $\mc{F}$ except that its only final state is~$s$.
  Then $\{ \mc{F}_s : s \in S(\mc{F})\}$ is a refinement of the
  partition $\{\mc{E}_i\}$ with the special property that for each
  automaton $\mc{F}_s$ and $a \in \Sigma(\mc{F})$, there is a unique
  automaton $\mc{F}_{s'}$ such that $\mc{F}_s \cdot a = \mc{F}_{s'}$.
  Indeed, since $\mc{F}$ is deterministic, $s'$ is the unique state
  corresponding to a transition $\trans{s}{a}{s'} \in T(\mc{F})$.
  
  If we let $\Gamma''(s) := \{ \mc{F}_{s'} : s' \in S(\mc{F}) \}$ for
  each state $s \in S(\mc{D})$, then we obtain finite refinements of
  $\Gamma(s)$ compatible with $\mc{D}$'s transition structure, as
  desired.
  
  This result implies that Algorithm~\ref{alg:optimize-domains} must
  eventually terminate.  After all, every refinement that
  Algorithm~\ref{alg:optimize-domains} performs must already be
  reflected in $\Gamma''(s)$.  Moreover, since every refinement that
  the algorithm performs is essential to compatibility with $\mc{D}$'s
  transition structure, the algorithm must, upon termination, produce
  the coarsest (smallest) compatible refinement possible.
\end{proof}

\begin{figure}[here]
  \includegraphics[width=2.6in]{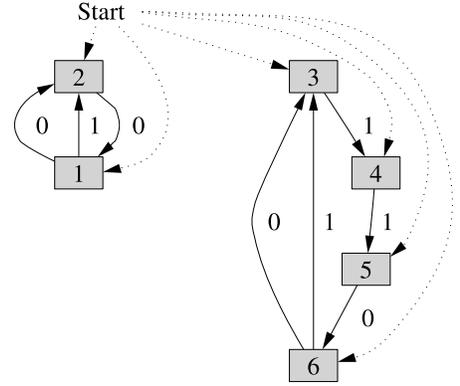}
  \caption{\label{fig:chaotic-domains} The positive-entropy domains
	$\mc{D}_1$ and $\mc{D}_2$ of the binary, next-to-nearest neighbor
    \CA~2614700074. (After Ref.~\cite{Crut93a}.)}
\end{figure}

\begin{figure}[here]
  \begin{minipage}[t]{2.4in}
    \includegraphics[width=2.4in]{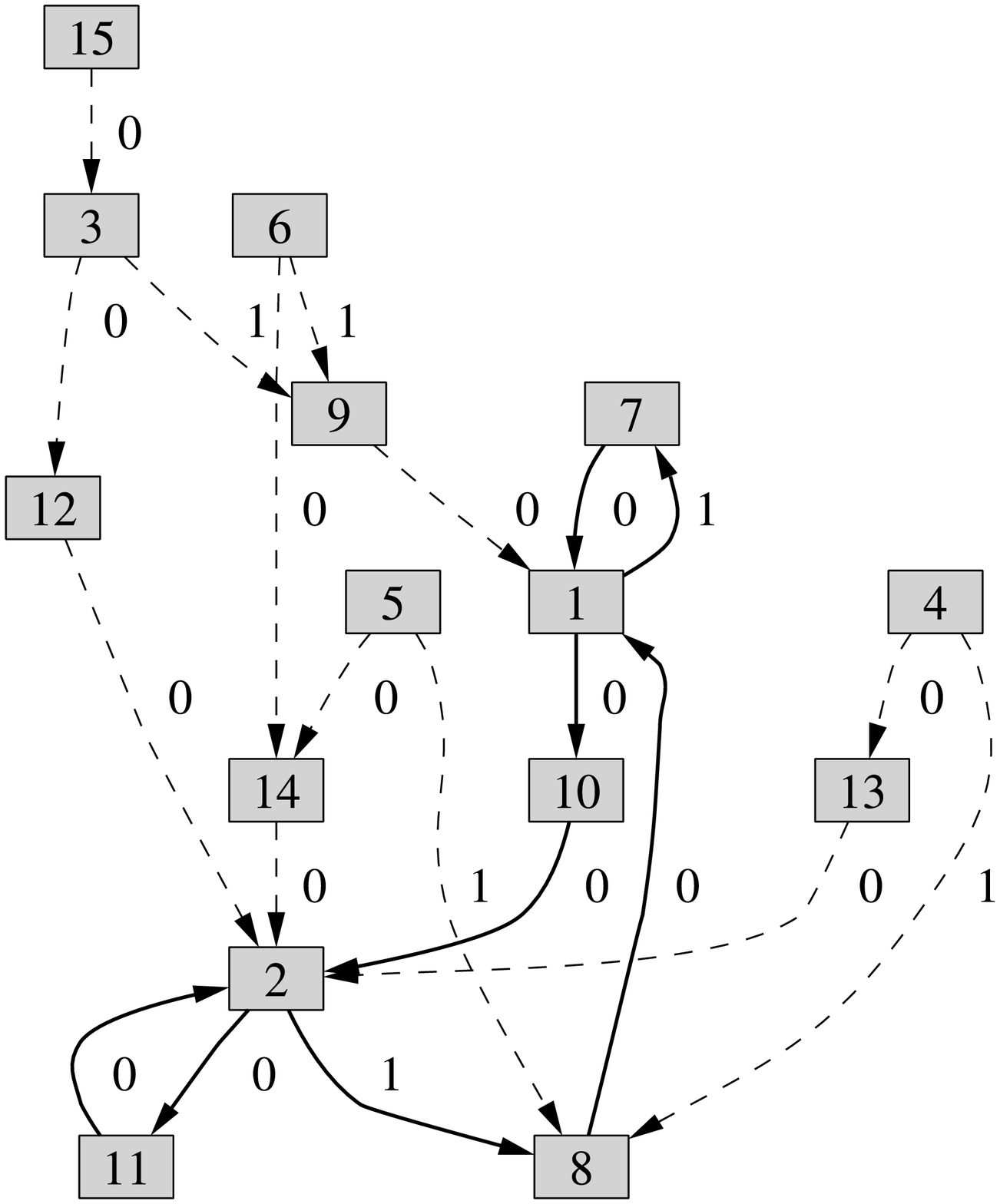}
  \end{minipage}
  \begin{minipage}[t]{2.5in}
    \includegraphics[width=2.5in]{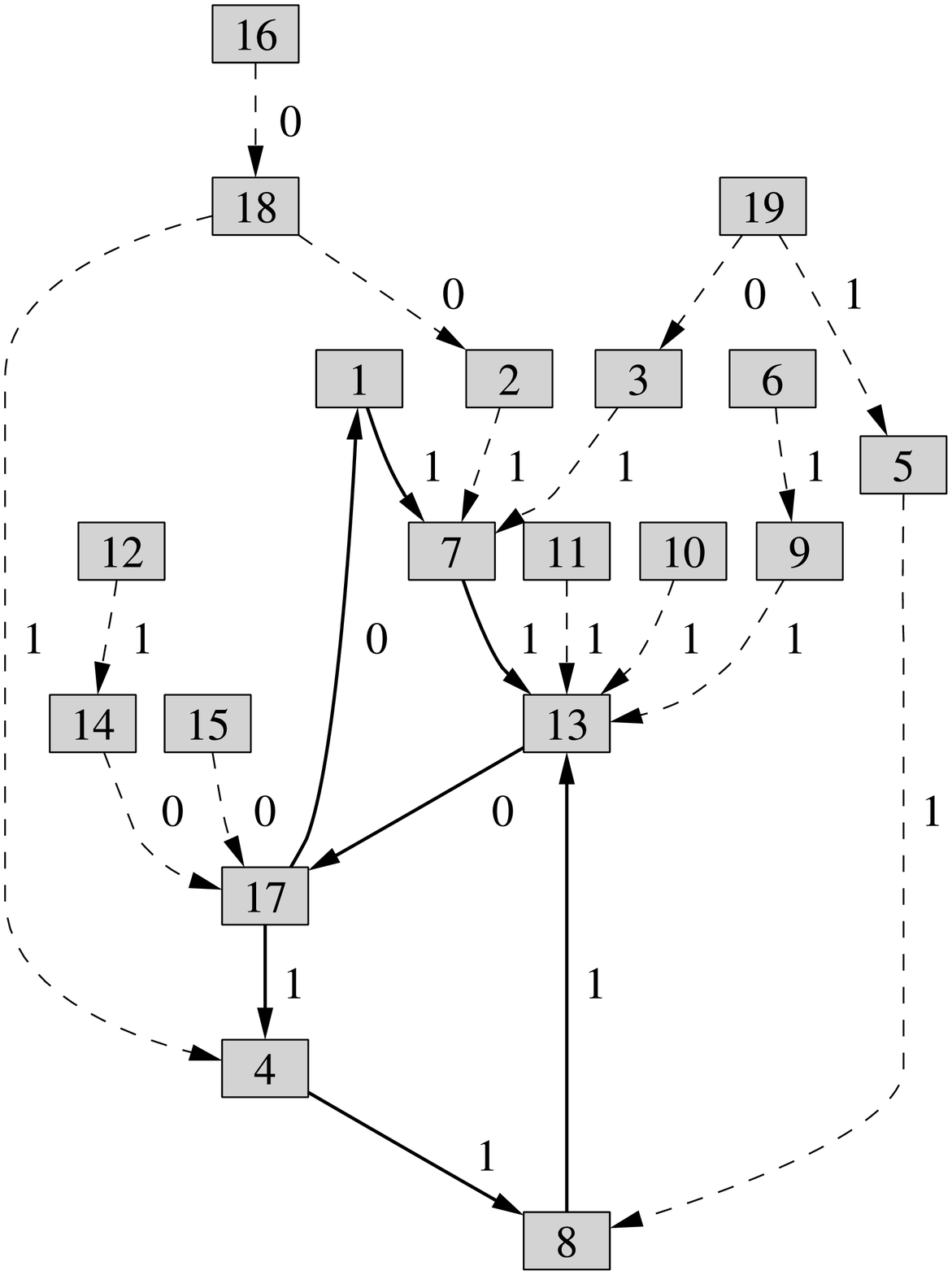}
  \end{minipage}
  \caption{\label{fig:optimized-chaotic-domains} The equivalent,
    non-minimal domains $\{\mc{D}'_1, \mc{D}'_2\} =
    \Optimize(\{\mc{D}_1, \mc{D}_2\})$ obtained by applying
    Algorithm~\ref{alg:optimize-domains} to the positive-entropy
    domains $\mc{D}_1$ and $\mc{D}_2$ in
    \Figure~\vref{fig:chaotic-domains}. ($\mc{D}'_1$ (top) and
	$\mc{D}'_2$ (bottom).) The ``Start'' arrows are
    omitted for clarity (all states are starting), and some of the
    transitions are drawn with dashed arrows to help the reader
    distinguish the recurrent states.}
\end{figure}

When applied to the domains $\mc{D}_1$ and $\mc{D}_2$ in
\Figure~\vref{fig:chaotic-domains}, for example,
Algorithm~\ref{alg:optimize-domains} produces the equivalent,
non-minimal domains $\{\mc{D}'_1, \mc{D}'_2\} = \Optimize(\{\mc{D}_1,
\mc{D}_2\})$ shown in \Figure~\vref{fig:optimized-chaotic-domains}.
Notice these domains' many non-recurrent states.  These have almost no
effect on the automaton $\mc{A}' := \Det(\bigsqcup_i \mc{D}'_i)$.

\section{Applications}

We now present four applications to illustrate how the stack-based
Algorithm~\ref{alg:stack-based} and its transducer approximation
(Algorithms~\ref{alg:transducer-building}
and~\ref{alg:optimize-domains}) solve the multi-regular language
filtering problem.  The first is the cellular automaton \ECA~110,
shown previously.  Its rather large filtering transducer is quite
tedious to construct by hand, but
Algorithm~\ref{alg:transducer-building} produces it handily.  The
second example, \ECA~18, which we have also already seen, illustrates
the stack-based Algorithm~\ref{alg:stack-based}'s ability to detect
overlapping domains.  The third example shows our methods' power to
detect structures in the midst of apparent randomness: the domains and
sharp boundaries between them are identified easily despite the fact
that the domains themselves have positive entropy and their boundaries
move stochastically.  The example shows the use of---and need
for---domain-preprocessing (Algorithm~\ref{alg:optimize-domains}).
That is, rapid resynchronization is achieved using a filter built from
optimized, non-minimal domains.  The final example demonstrates the
transducer (constructed by Algorithms~\ref{alg:transducer-building}
and~\ref{alg:optimize-domains}) detecting domains in a
multi-stationary process---what is called the change-point problem in
statistical time-series analysis.  This example emphasizes that the
methods developed here are not limited to cellular automata. More
importantly, it highlights several of the subtleties of multi-regular
language filtering and clearly illustrates the need for the
domain-preprocessing Algorithm~\ref{alg:optimize-domains}.

\subsection*{\CapitalECA~110}

\begin{figure}
  \includegraphics[width=3in]{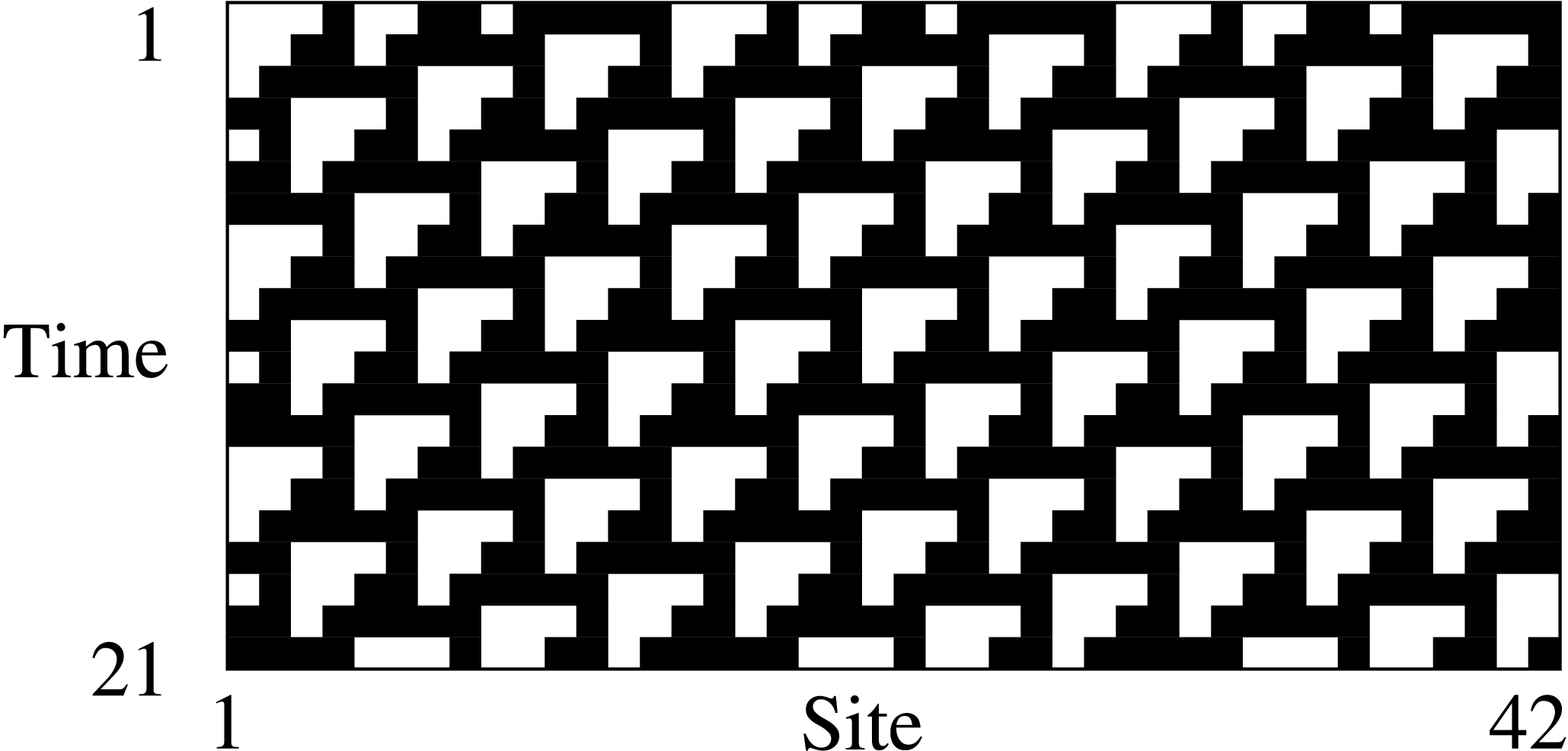}
  \caption{\label{fig:eca110-spacetime-domain} \ECA~110's  principal domain,
    $\sub[(00010011011111)^*]$.}
\end{figure}

\begin{figure*}
\includegraphics[scale=0.3]{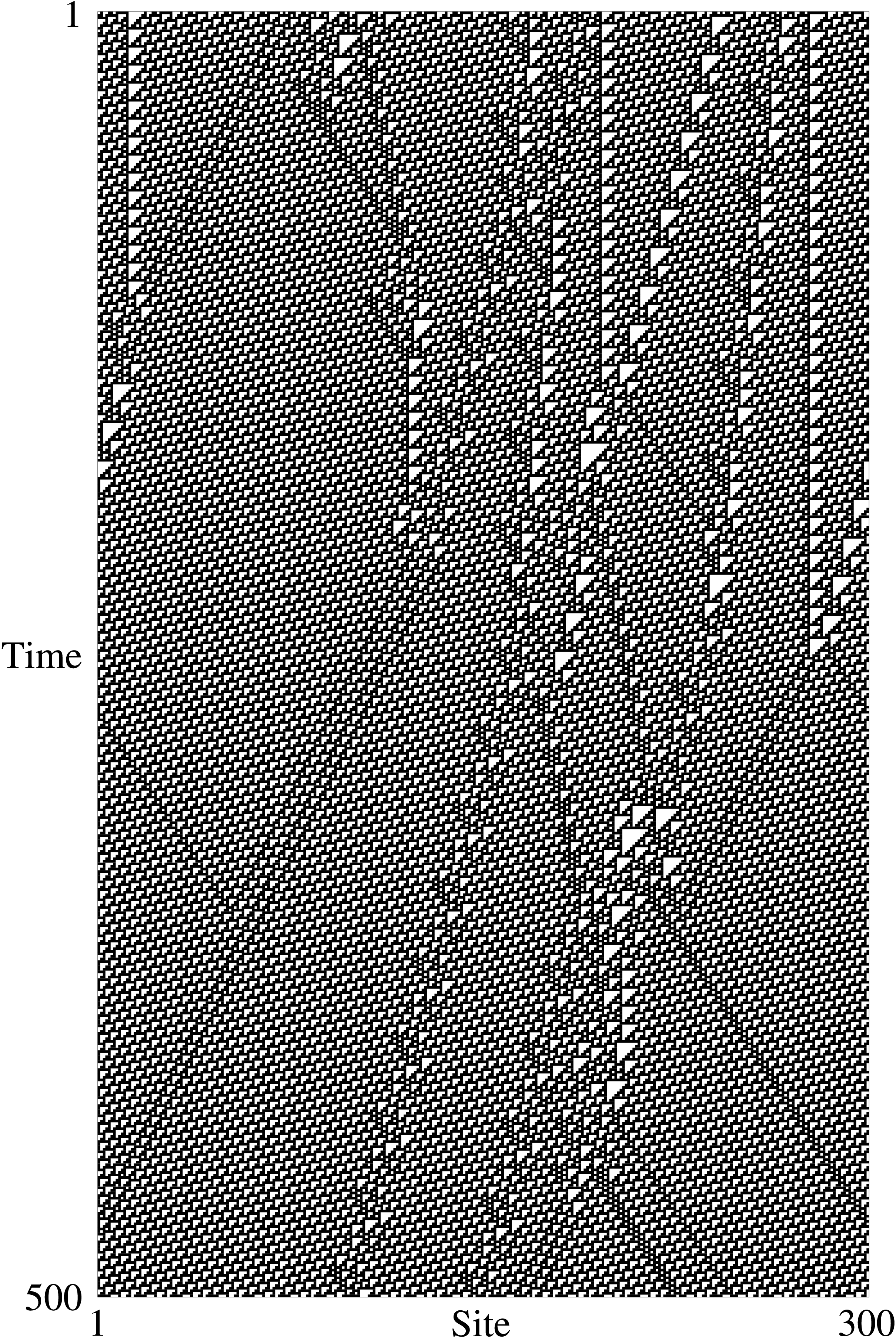}
\includegraphics[scale=0.3]{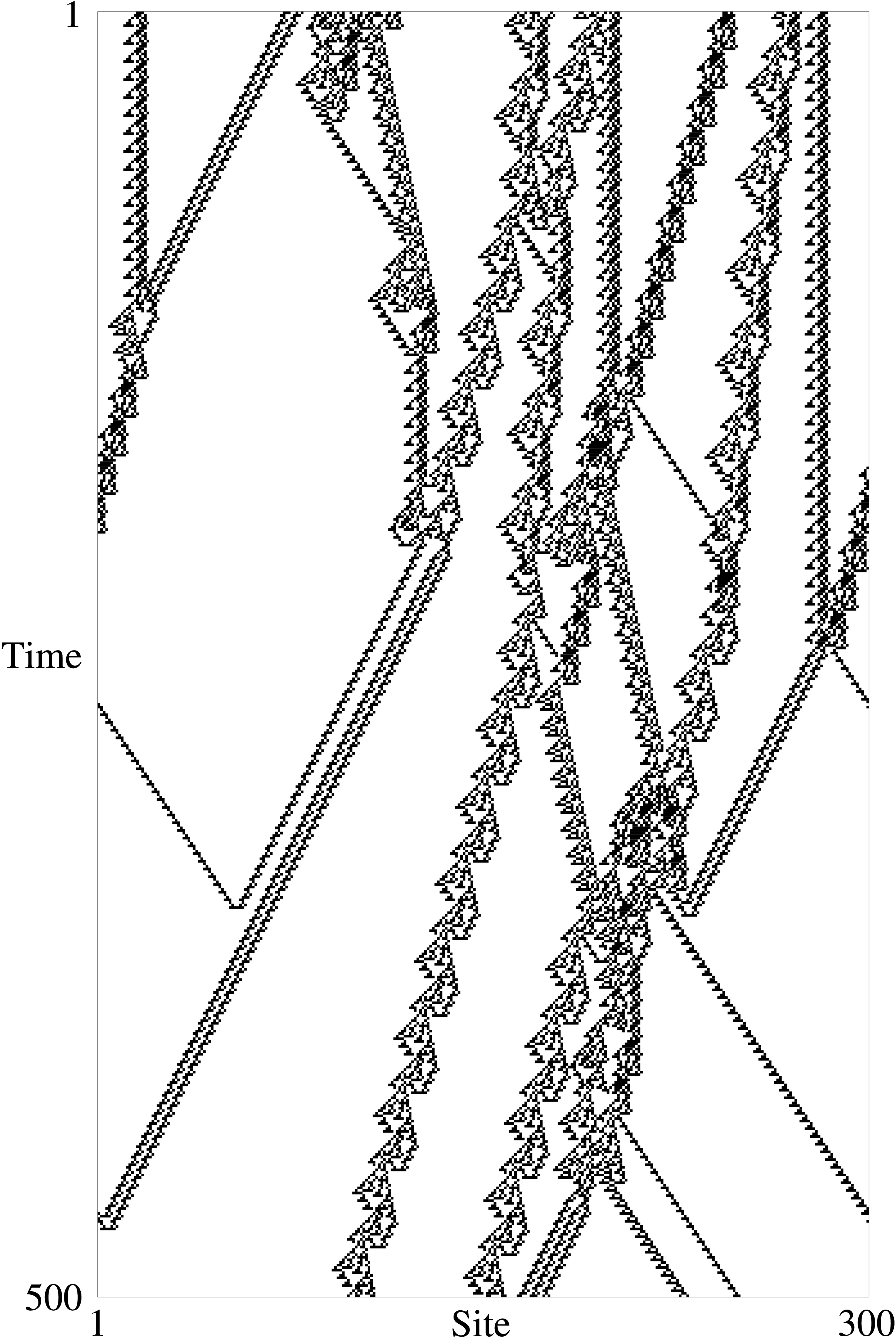}
\caption{An \ECA~110 space-time diagram (left) filtered by the transducer
  $\Filter(\{\sub[(00010011011111)^*]\})$ (right).
  \label{eca110-big}}
\end{figure*}

First consider \ECA~110, illustrated earlier in \Figure~\vref{eca110}.
Its domains are easy to see visually; they have the form $\sub(w^*)$
for some finite word~$w$.  Its dominant domain is $\sub(w^*) =
\sub[(00010011011111)^*]$, illustrated in
\Figure~\vref{fig:eca110-spacetime-domain}.
In fact, the transducer $\Filter(\{ \sub[(00010011011111)^*] \})$,
constructed from this single domain, filters \ECA~110's space-time
behavior well; see \Figure~\vref{eca110-big}.

Notice, in that figure, the wide variety of particle-like domain
defects that the filtered version lays bare.  Note, moreover, how
these particles move and collide according to consistent rules.  These
particles are important to \ECA~110's computational properties; a
subset can be used to implement a Post Tag system \cite{Mins67} and
thus simulate arbitrary Turing machines \cite{Cook04}.

\subsection*{\CapitalECA~18}

Next, consider \CapitalECA~18, illustrated earlier in
\Figure~\vref{eca18}.  It is somewhat more challenging to filter,
because its domain $\mc{D}=\sub \left( [0(0+1)]^* \right)$ has
positive entropy.  As a result, its particles are
difficult---although by no means impossible---to see with the naked
eye.  Nevertheless, the stack-based algorithm filters its space-time
diagrams extremely well, as illustrated in \Figure~\ref{eca18}
(right).  There, black rectangles are drawn where maximal substrings
overlap, and vertical bars are drawn where maximal substrings abut.
As mentioned earlier, these particles, whose precise location is
somewhat ambiguous, follow random walks and pairwise annihilate
whenever they touch~\cite{Crut92a,Eloranta94,Hans90a}.

It is worth mentioning that the transducer $\Filter(\{\mc{D}\})$
produces a less precise filtrate in this case---and that
$\Filter(\Optimize(\{\mc{D}\}))$ does no better.  Indeed, since breaks
in \ECA~18's domain have the form $\cdots 1 (0^{2 n}) 1 \cdots$, the
precise location of the domain break is ambiguous: if reading
left-to-right, it does not occur until the $1$ on the right of $0^{2
  n}$ is read; whereas, if reading right-to-left, it does not occur
until the $1$ on the left is read.  In other words, if reading
left-to-right, the transducer $\Filter(\{\mc{D}\})$ detects only the
right edges of the black triangles of \Figure~\ref{eca18} (right).
Similarly, if reading right-to-left, it detects only the left edges of
these triangles.  In this case it is possible to fill in the space
between these pairs of edges to obtain the output of the stack-based
algorithm.

\subsection*{\CapitalCA~2614700074}

\begin{figure*}
\includegraphics[scale=0.3]{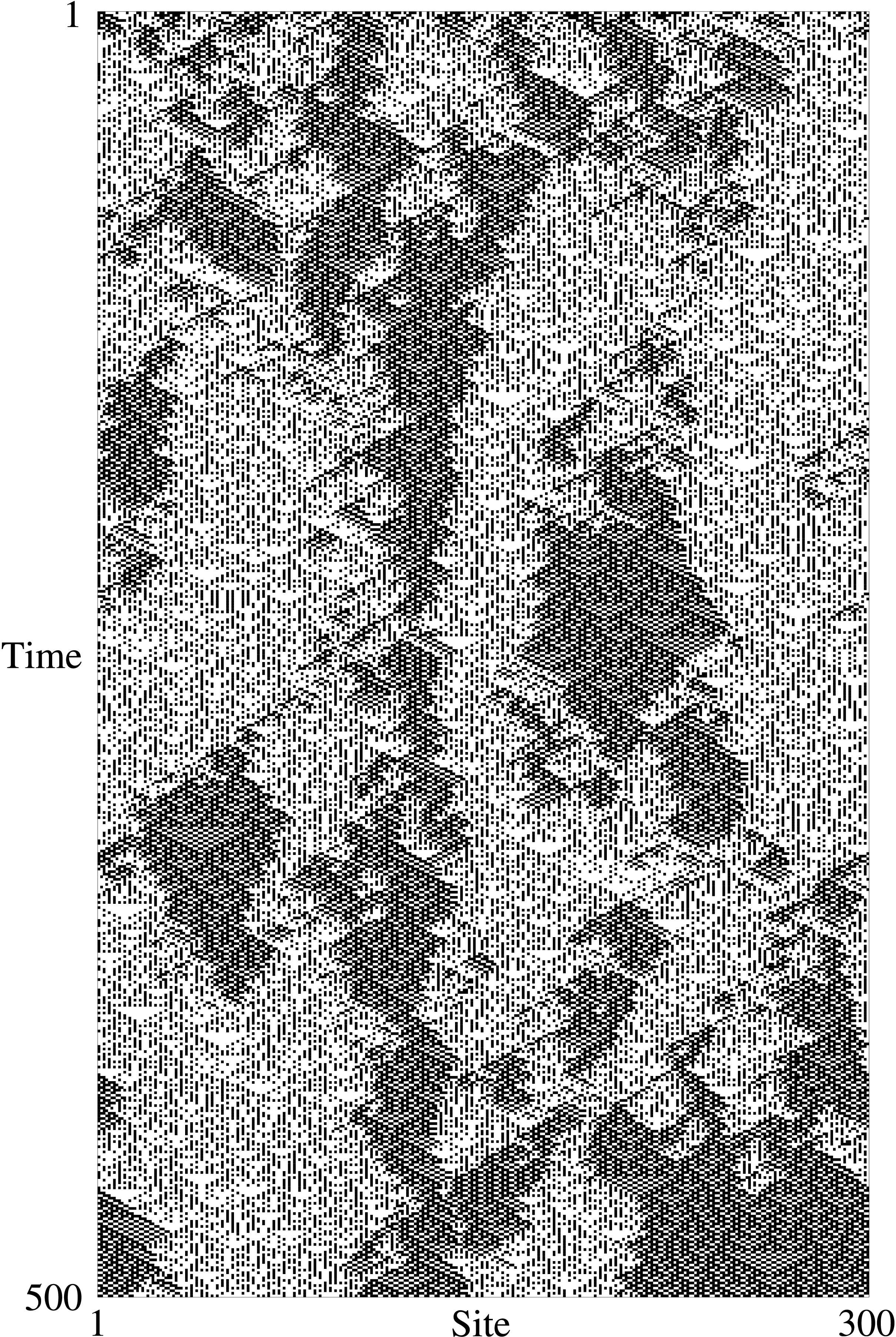}
\includegraphics[scale=0.3]{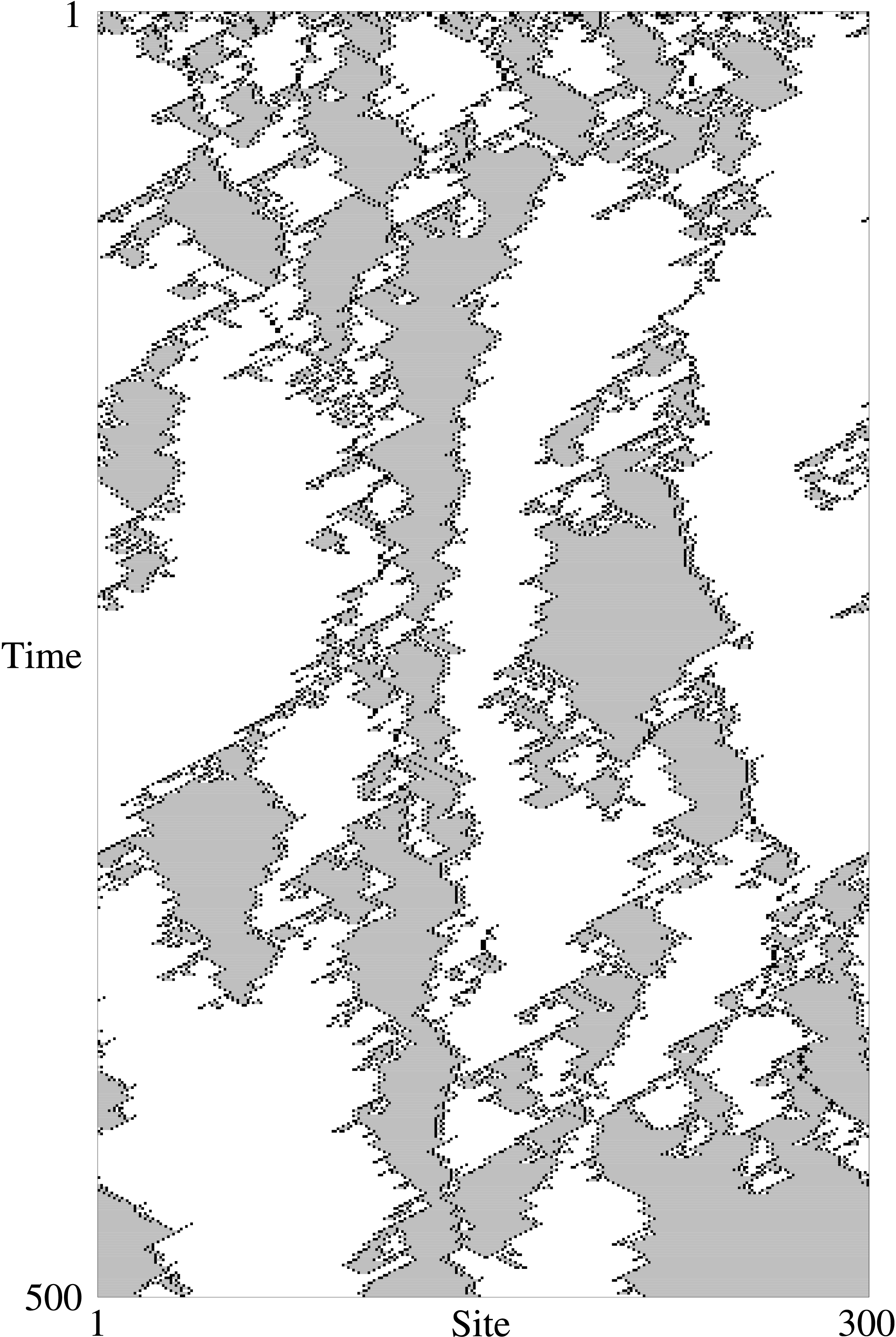}
\caption{Binary, next-to-nearest neighbor \CA~2614700074 space-time
  diagram (left) filtered by the transducer
  $\Filter(\Optimize(\{\mc{D}_1,\mc{D}_2\}))$ (right).  The white
  regions on the right correspond to the domain $\mc{D}_1$, the gray
  to the domain $\mc{D}_2$.  The black squares separating these
  regions correspond to the interruption symbols $h'(s,s')$ that the
  transducer emits between domains.
  \label{chaotic-ca-big}}
\end{figure*}

Now consider the binary, next-to-nearest neighbor (i.e. $k$=$r$=2)
\CA~2614700074, shown in \Figure~\vref{chaotic-ca-big}.  Crutchfield
and Hanson constructed it expressly to have the positive-entropy domains
$\mc{D}_1$ and $\mc{D}_2$ in \Figure~\vref{fig:chaotic-domains} \cite{Crut93a}.

As illustrated in \Figure~\vref{chaotic-ca-big}, the optimal
transducer $\Filter(\Optimize(\{\mc{D}_1,\mc{D}_2\}))$ filters this
\CA's output well.  This illustrates a practical advantage of
multi-regular language filtering: it can detect structure embedded in
randomness.  Notice how the filter easily identifies the domains and
sharp boundaries separating them, even though the domains themselves
have positive entropy and their boundaries move stochastically.

It is worth noting that in place of the gray regions of
\Figure~\ref{chaotic-ca-big} so clearly identified by the optimal
transducer as corresponding to the second domain~$\mc{D}_2$, the
simpler transducer $\Filter(\{\mc{D}_1,\mc{D}_2\})$ produces a regular
checkering of false domain breaks (not pictured).  This is because,
when examining the sole forbidden transition $(s,a) = (2,1)$ of the
first domain~$\mc{D}_1$, Algorithm~\ref{alg:transducer-building}
discovers that the first non-empty set $S_{i=1,l=4} = \{2,4,5\}$
contains three resynchronization states.  It unfortunately abandons
both states~4 and~5, which belong to the second domain, instead
choosing to resynchronize to the original state~2 itself, because it
occurs alone in the next set $S_{1,5}$.  As a result, the transducer
$\Filter(\{\mc{D}_1,\mc{D}_2\})$ has no transitions leaving the first
domain whatsoever and is therefore incapable of detecting jumps from
the first domain to the second.  This is why it prints a checkering of
domain breaks instead of correctly resynchronizing to the second
domain.  The optimal transducer does not suffer from this problem,
because Algorithm~\ref{alg:optimize-domains} splits state~2 into
several new ones, from which unambiguous resynchronization to the
appropriate state---2, 4, or 5---is possible.

\subsection*{Change-Point Problem: Filtering Multi-Stationary Sources}

\begin{figure}[here]
  \begin{minipage}[t]{2.4in}
    \includegraphics[width=1.9in]{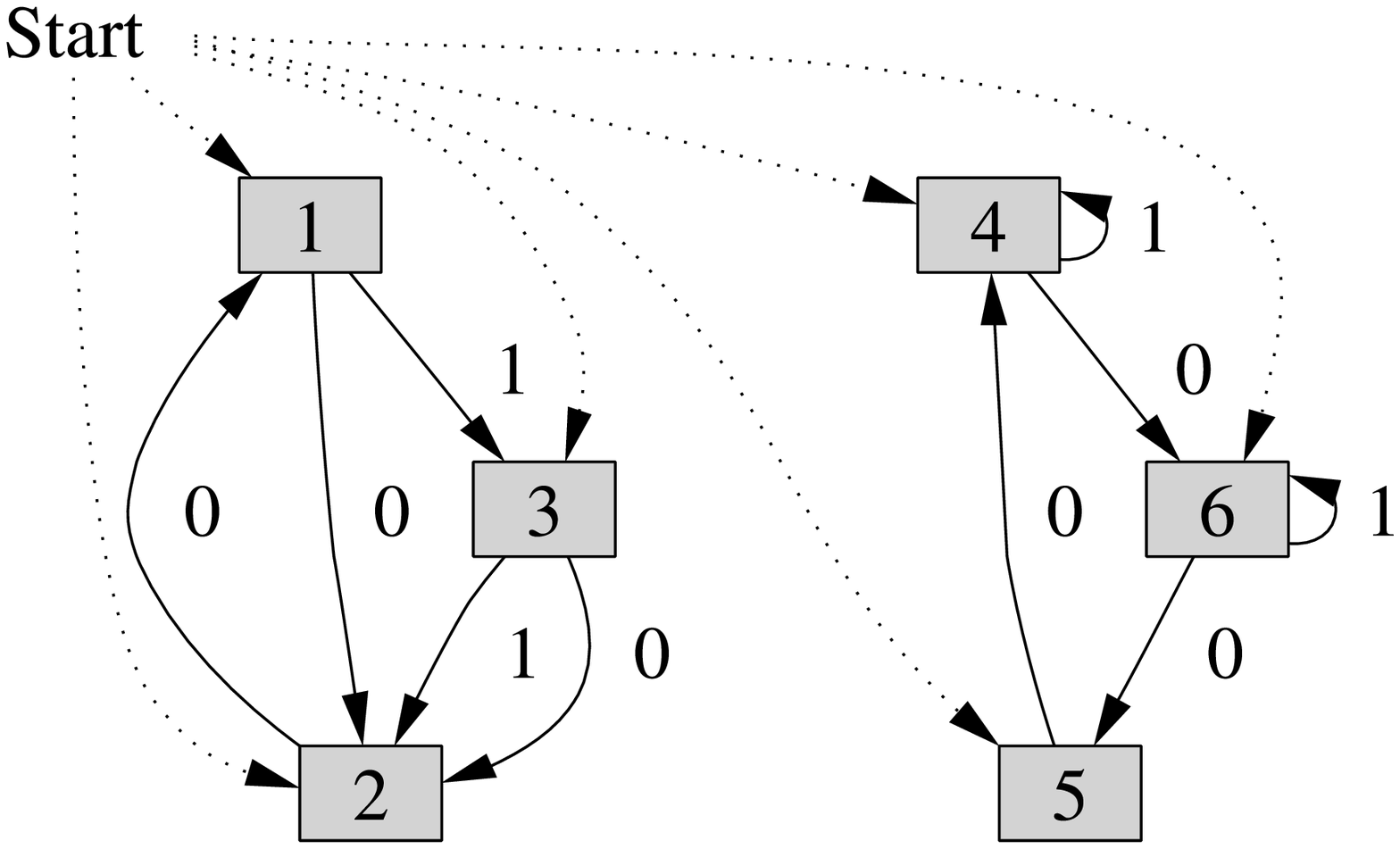}
  \end{minipage}
  \begin{minipage}[t]{2.3in}
    \includegraphics[width=2.5in]{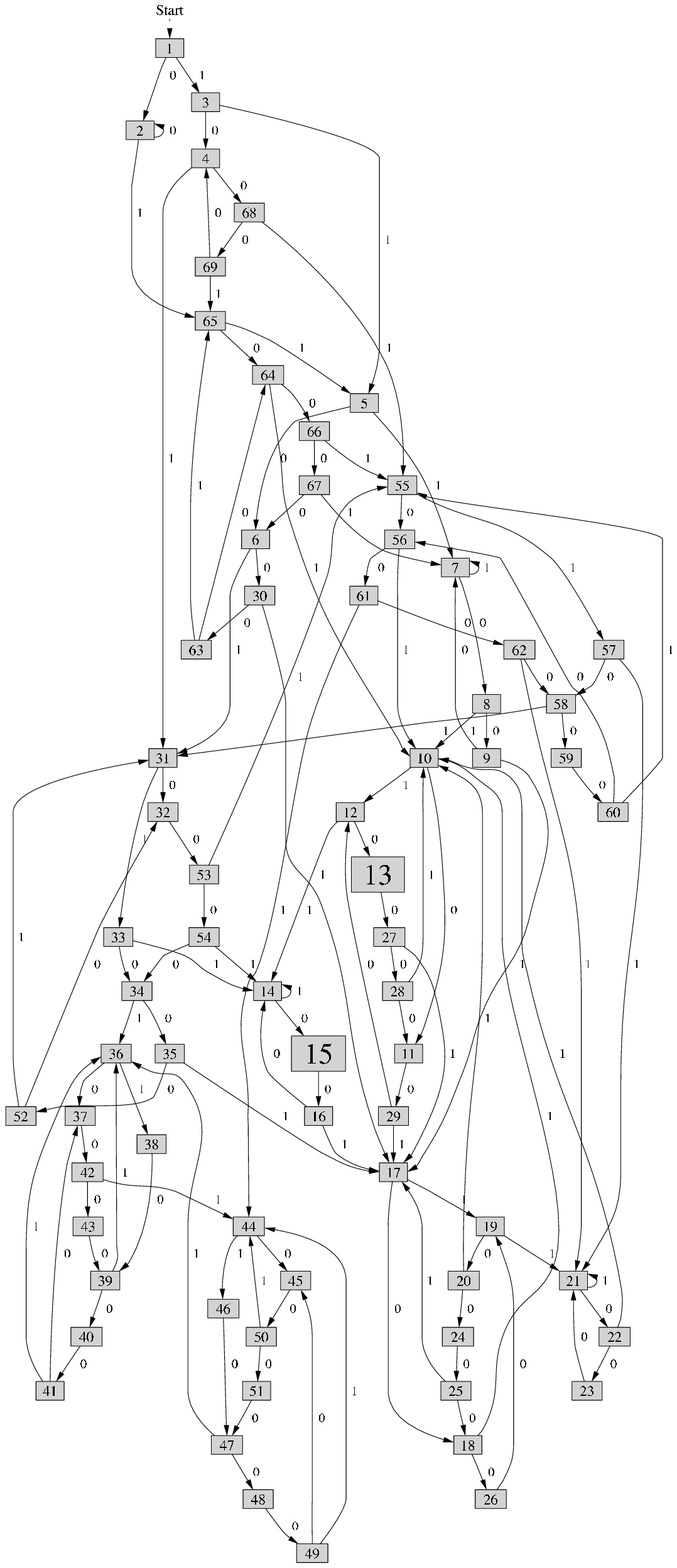}
  \end{minipage}
  \caption{\label{fig:nasty-domains}Two similar three-state domains
    $\mc{D}_1$ (top left) and $\mc{D}_2$ (top right) illustrate how
	subtle the construction of the optimal transducer
    $\Filter(\Optimize(\{\mc{D}_i\}))$ can be: the automaton $\mc{A}'
    := \Det(\bigsqcup \Optimize(\{\mc{D}_1, \mc{D}_2\}))$ (below),
    from which the optimal transducer is constructed, has 69
    states---the unoptimized automaton $\mc{A} := \Det(\mc{D}_1 \sqcup
    \mc{D}_2)$ (not pictured) has 30.  }
\end{figure}

Leaving cellular automata behind, consider a binary information source
that hops with low probability between the two three-state domains
$\mc{D}_1$ and $\mc{D}_2$ in \Figure~\vref{fig:nasty-domains} (top).
This source allows us to illustrate subtleties in multi-regular
language filtering and, in particular, in the construction of the
optimal transducer $\Filter(\Optimize(\{\mc{D}_i\})$ can be.

To appreciate how subtle filtering with the domains $\mc{D}_1$ and
$\mc{D}_2$ is---and why the extra states of
$\Optimize(\{\mc{D}_1,\mc{D}_2\})$ are needed to do it---consider the
following.  First choose any finite word $w$ of the form:
\begin{align*}
  (0^6 + 0^3 1^2)^* 0^3 1^2 0.
\end{align*}

As the ambitious reader can verify, both of the strings $101111w$ and
$110w$ belong to the domain $\mc{D}_2$.  In fact, both correspond to
unique paths through $\mc{D}_1 \sqcup \mc{D}_2$ ending in state~5 of
\Figure~\vref{fig:nasty-domains} (top).

On the other hand, the strings $01111w1$ and $10w1$ are also domain
words---the first belonging to $\mc{D}_2$, but the second belonging to
$\mc{D}_1$.  In fact, $01111w1$ corresponds to a unique path through
$\mc{D}_1 \sqcup \mc{D}_2$ ending in \emph{state~6}, while $10w1$
corresponds to a unique path ending in \emph{state~3}.

As a result, these four strings are the maximal substrings of the
non-domain strings $101111w1$ and $110w1$, as indicated by the
brackets below:

\begin{align*}
  \overunderbraces{&\br{2}{\parbox{1.7in}{\small corresponds to a
        unique path through $\mc{D}_2$ ending in state 5}}&}
  {&1\;&0\;1\;1\;1\;1\;w&\;1&} {&&\br{2}{\text{\parbox{1.7in}{\small
          corresponds to a unique path through $\mc{D}_2$ ending in
          \emph{state 6}}}}}
\end{align*}
\begin{align*}
  \overunderbraces{&\br{2}{\text{\parbox{1.7in}{\small corresponds to
          a unique path through $\mc{D}_2$ ending in state 5}}}&}
  {&1\;&1\;0\;w&\;1&} {&&\br{2}{\text{\parbox{1.7in}{\small
          corresponds to a unique path through $\mc{D}_1$ ending in
          \emph{state 3}}}}}
\end{align*}

This example illustrates several important points.  First of all, it
shows that when the naive transducer $\Filter(\{\mc{D}_1, \mc{D}_2\})$
reaches the forbidden letter 1 at the end of either of these two
strings, the state~2 reached does not preserve enough information to
resynchronize to the appropriate state---3 or 6, respectively.  As a
result, it must either make a guess---at the risk of choosing incorrectly
and then later reporting an artificial domain break (as in the preceding
cellular automaton example)---or else jump to one of its non-recurrent
states, emitting a potentially long chain of $\lambda$s until it can
re-infer from future input what was already determined by past input.

As unsettling as this may be, the example illustrates something far
more nefarious. Since an arbitrarily long word $w$ can be chosen, it
is impossible to fix the problem by splitting the states of
$\Filter(\{\mc{D}_1,\mc{D}_2\})$ so as to buffer finite windows of
past input.  In fact, because $w$ is chosen from a language with
positive entropy, the number of windows that would need to be buffered
grows exponentially.

At this point achieving optimal resynchronization might seem hopeless,
but it actually is possible. This is what makes
Algorithm~\ref{alg:optimize-domains}---and in particular the proof
that it terminates
(Prop.~\ref{prop:optimization-terminates})---not only
surprising, but extremely useful.

Indeed, recall that instead of splitting states according to finite
windows, Algorithm~\ref{alg:optimize-domains} splits them according to
entire regular languages of past input and that, by
Prop.~\ref{prop:optimization-terminates}, a finite number of
these regular languages will always suffice to achieve optimal
resynchronization.  And so, instead of reaching the same original
state~2 when reading the strings $101111w$ and $110w$, the optimal
transducer $\Filter(\Optimize(\{\mc{D}_1,\mc{D}_2\}))$ reaches two
distinct states $(2,\mc{E})$ and $(2,\mc{E}')$, where $101111w \in
\Lang(\mc{E})$ and $110w \in \Lang(\mc{E}')$.  These two split states
are labeled with the enlarged integers 15 and 13, respectively, in
\Figure~\ref{fig:nasty-domains} (bottom), which shows $\mc{A}' :=
\Det(\bigsqcup \Optimize(\{\mc{D}_1, \mc{D}_2\}))$---the automaton
from which $\Filter(\Optimize(\{\mc{D}_1,\mc{D}_2\}))$ is constructed.
As illustrated in that figure, the optimal transducer has $69$
states---the unoptimized automaton $\mc{A} := \Det(\mc{D}_1 \sqcup
\mc{D}_2)$ (not pictured) has $30$.

\section{Conclusion}

We posed the multi-regular language filtering problem and presented
two methods for solving it.  The first, although providing the ideal
solution, requires a stack, has a worst-case compute time that grows
quadratically in string length and conditions its output at any
point on arbitrarily long windows of future input.  The second method
was to algorithmically construct a transducer that approximates the first
algorithm.  In contrast to the stack-based algorithm it approximates,
however, the transducer requires only a finite amount of memory, runs
in linear time, and gives immediate output for each letter
read---significant improvements for cellular automata structural
analysis and, we suspect, for other applications as well.  It is,
moreover, the best possible approximation with these three features.
Finally, we applied both methods to the computational-mechanics
structural analysis of cellular automata and to a version of the
change-point problem from time-series analysis.

Future directions for this work include generalization both to
probabilistic patterns and transducers and to higher dimensions.
Although both seem difficult, the latter seems most daunting---at
least from the standpoint of transducer construction---because there
is as yet no consensus on how to approach the subtleties of
high-dimensional automata theory.  (See, for example,
Refs.~\cite{Feld02b} and \cite{Lindgren98} for discussions of
two-dimensional generalizations of regular languages and patterns.)
Note, however, that the basic notion of maximal substrings underlying
the stack-based algorithm is easily generalized to a broader notion of
higher-dimensional maximal connected subregions, although we suspect
that this generalization will be much more difficult to compute.

In the introduction we alluded to a range of additional applications
of multi-regular language filtering.  Segmenting time series into
structural components was illustrated by the change-point example.
This type of time series problem occurs in many areas, however, such
as in speech processing where the structural components are hidden
Markov models of phonemes, for example, and in image segmentation
where the structural components are objects or even textures.  One of
the more promising areas, though, is genomics.  In genomics there is
often quite a bit of prior biochemical knowledge about structural
regions in biosequences. Finally, when coupled with statistical
inference of stationary domains, so that the structural components are
estimated from a data stream, multi-regular language filtering should
provide a powerful and broadly applicable pattern detection tool.

\section*{Acknowledgments}

This work was supported at the Santa Fe Institute under the
Networks Dynamics Program funded by the Intel Corporation and under
the Computation, Dynamics, and Inference Program via SFI's core
grants from the National Science and MacArthur Foundations. Direct
support was provided by DARPA Agreement F30602-00-2-0583.

\appendix

\section{Automata Theory Preliminaries}

\label{automata-review}

In this appendix we review the definitions and results from automata
theory that are essential to our exposition.  A good source for these
preliminaries is Ref.~\cite{Hopc79}, although its authors employ
altogether different notation, which does not suit our needs.

\subsection*{Automata}

An \emph{automaton} $\mc{A}$ over an alphabet~$\Sigma(\mc{A})$ is a
collection of states $S(\mc{A})$, together with subsets
$\mathrm{Start}(\mc{A}), \mathrm{Final}(\mc{A}) \subset S(\mc{A})$,
and a collection of transitions $T(\mc{A}) \subset S(\mc{A}) \times
\Sigma(\mc{A}) \times S(\mc{A})$.  We call an automaton \emph{finite}
if both $S(\mc{A})$ and $T(\mc{A})$ are.

An automaton $\mc{A}$ \emph{accepts} a string $\sigma = a_1 a_2 \cdots
a_n$ if there is a sequence of transitions $\trans{s_1}{a_1}{s_2},
\trans{s_2}{a_2}{s_3}, \dots, \trans{s_{n-1}}{a_n}{s_n} \in T(\mc{A})$
such that $s_1 \in \mathrm{Start}(\mc{A})$ and $s_n \in
\mathrm{Final}(\mc{A})$.  Denote the collection of all strings that
$\mc{A}$ accepts by $\Lang(\mc{A})$.  Two automata $\mc{A}$ and
$\mc{B}$ are said to be \emph{equivalent} if
$\Lang(\mc{A})=\Lang(\mc{B})$.

We can think of an automaton as a directed graph whose edges are
labeled with symbols from $\Sigma(\mc{A})$.  In this view, an
automaton accepts precisely those strings that correspond to paths
through its graph beginning in its start states and ending in its
final ones.

An automaton $\mc{A}$ is said to be \emph{semi-deterministic} if any
pair of its transitions that agree in the first two slots are
identical, that is, any pair of transitions of the form
$\trans{s_1}{a}{s_2}$ and $(s_1,a,s'_2) \in T(\mc{A})$ satisfy $s_2 =
s'_2$.  A \emph{deterministic} automaton is one that is
semi-deterministic and that has a single start state.  If $\mc{A}$ is
deterministic, then each string of $\Lang(\mc{A})$ corresponds to
precisely one path through $\mc{A}$'s graph.

For two automata $\mc{A}$ and $\mc{B}$, let $\mc{A} \sqcup \mc{B}$
denote their \emph{disjoint union}---the automaton over the alphabet
$\Sigma(\mc{A}) \cup \Sigma(\mc{B})$ whose states are the disjoint
union of the states of $\mc{A}$ and $\mc{B}$, i.e. $S(\mc{A} \sqcup
\mc{B}) = S(\mc{A}) \sqcup S(\mc{B})$ (and similarly for its start
and final states) and whose transitions are the union of the
transitions of $\mc{A}$ and $\mc{B}$.  In this way, $\Lang(\mc{A}
\sqcup \mc{B}) = \Lang(\mc{A}) \cup \Lang(\mc{B})$.

In this terminology, a \emph{domain} is a semi-deterministic finite
automaton $\mc{D}$ whose states are all start and final states, i.e.
$\mathrm{Start}(\mc{D}) = S(\mc{D}) = \mathrm{Final}(\mc{D})$, and
whose graph is strongly connected---i.e., there is a path from any
one state to any other.

Finally, a domain $\mc{D}$ is said to be \emph{minimal} if all
equivalent domains $\mc{D}'$ satisfy $|S(\mc{D})| \le |S(\mc{D}')|$.

\subsection*{Standard Results}

\begin{lem}
  \label{subset-construction}
  Every automaton $\mc{A}$ is equivalent to a deterministic automaton
  $\Det(\mc{A})$.  Moreover, $\Det(\mc{A})$'s states correspond
  uniquely to collections of $\mc{A}$'s states; in other words, there
  is a canonical injection $S(\Det(\mc{A}))
  \stackrel{\psi_{\Det(\mc{A})}}{\hookrightarrow} \{ S : S \subset
  S(\mc{A}) \}$.
\end{lem}


\begin{lem}
  \label{intersection-construction}
  If $\mc{A}$ and $\mc{B}$ are automata, then there is an automaton
  $\mc{A} \cap \mc{B}$ that accepts precisely those strings accepted
  by both $\mc{A}$ and $\mc{B}$; that is, $\Lang(\mc{A} \cap \mc{B}) =
  \Lang(\mc{A}) \cap \Lang(\mc{B})$.  If $\mc{A}$ and $\mc{B}$ are
  deterministic, then so is $\mc{A} \cap \mc{B}$.  Moreover, there is
  a canonical injection $S(\mc{A} \cap \mc{B}) \hookrightarrow
  S(\mc{A}) \times S(\mc{B})$, which restricts to injections
  $\mathrm{Start}(\mc{A} \cap \mc{B}) \hookrightarrow
  \mathrm{Start}(\mc{A}) \times \mathrm{Start}(\mc{B})$ and
  $\mathrm{Final}(\mc{A} \cap \mc{B}) \hookrightarrow
  \mathrm{Final}(\mc{A}) \times \mathrm{Final}(\mc{B})$.
\end{lem}


\subsection*{Transducers}

A \emph{transducer} $\mc{T}$ from an alphabet $\Sigma(\mc{T})$ to an
alphabet $\Sigma'(\mc{T})$ is an automaton on the alphabet
$\Sigma(\mc{T}) \times \Sigma'(\mc{T})$.  We will use the more
traditional notation $(s,b|c,s')$ in place of $(s,(b,c),s') \in
T(\mc{T})$.

The \emph{input} of a transducer $\mc{T}$ is the automaton
$\Input(\mc{T})$ whose states, start states, and final states are the
same as $\mc{T}$'s, but whose transitions are given by
$T(\Input(\mc{T})) := \{ \trans{s}{b}{s'} : \trans{s}{b|c}{s'} \in
T(\mc{T}) \}$.  Similarly, the \emph{output} of a transducer $\mc{T}$
is the automaton $\Output(\mc{T})$ whose transitions are given by
$T(\Output(\mc{T})) := \{ \trans{s}{c}{s'} : \trans{s}{b|c}{s'} \in
T(\mc{T}) \}$.

A transducer $\mc{T}$ is said to be \emph{well defined} if
$\Input(\mc{T})$ is deterministic, because such a transducer
determines a function from $\Lang(\Input(\mc{T}))$ onto
$\Lang(\Output(\mc{T}))$.


\label{automata-review-end}

\section{Implementation}

\label{implementation}


In order to give the reader a sense for how the algorithms
can be implemented, we rigorously implement
Algorithm~\ref{alg:transducer-building} here in the programming
language Haskell~\cite{Peyton03}.  Haskell represents the state of the
art in polymorphicly typed, lazy, purely functional programming
language design.  Its concise syntax enables us to implement the
algorithm in less than a page.  Haskell compilers and interpreters are
freely available for almost any computer~\footnote{For Haskell
compilers and interpreters, please visit \url{www.haskell.org}.}.

We emulate our exposition in the preceding sections by representing a
finite automaton as a list of starting states, a list of transitions,
and a list of final states, and a transducer as a finite automaton
whose alphabet consists of pairs of symbols:

\begin{tabbing}
\qquad\=\hspace{\lwidth}\=\hspace{\cwidth}\=\+\kill
${\mathbf{data}\;\Conid{FA}\;\Varid{s}\;\Varid{i}\mathrel{=}\Conid{FA}\{\mskip1.5mu \Varid{faStarts}\mathbin{::}[\mskip1.5mu \Varid{s}\mskip1.5mu],}$\\
${\phantom{\mathbf{data}\;\Conid{FA}\;\Varid{s}\;\Varid{i}\mathrel{=}\Conid{FA}\{\mskip1.5mu \mbox{}}\Varid{faTrans}\mathbin{::}[\mskip1.5mu (\Varid{s},\Varid{i},\Varid{s})\mskip1.5mu],}$\\
${\phantom{\mathbf{data}\;\Conid{FA}\;\Varid{s}\;\Varid{i}\mathrel{=}\Conid{FA}\{\mskip1.5mu \mbox{}}\Varid{faFinals}\mathbin{::}[\mskip1.5mu \Varid{s}\mskip1.5mu]\mskip1.5mu\}}$\\
${}$\\
${\mathbf{type}\;\Conid{Transducer}\;\Varid{s}\;\Varid{i}\;\Varid{o}\mathrel{=}\Conid{FA}\;\Varid{s}\;(\Varid{i},\Varid{o})}$
\end{tabbing}

We need the following simple functions, which compute the list of
symbols and states present in an automaton:
\begin{tabbing}
\qquad\=\hspace{\lwidth}\=\hspace{\cwidth}\=\+\kill
${\Varid{faAlphabet}\mathbin{::}\Conid{Eq}\;\Varid{i}\Rightarrow \Conid{FA}\;\Varid{s}\;\Varid{i}\to [\mskip1.5mu \Varid{i}\mskip1.5mu]}$\\
${\Varid{faAlphabet}\;\Varid{fa}\mathrel{=}\Varid{nub}\;[\mskip1.5mu \Varid{a}\mid (\anonymous ,\Varid{a},\anonymous )\leftarrow \Varid{faTrans}\;\Varid{fa}\mskip1.5mu]}$\\
${}$\\
${\Varid{transStates}\mathbin{::}\Conid{Eq}\;\Varid{s}\Rightarrow [\mskip1.5mu (\Varid{s},\Varid{i},\Varid{s})\mskip1.5mu]\to [\mskip1.5mu \Varid{s}\mskip1.5mu]}$\\
${\Varid{transStates}\;\Varid{trans}\mathrel{=}}$\\
${\hskip2.00em\relax\Varid{nub}\mathbin{\$}\Varid{foldl}\;(\lambda \Varid{ss}\;(\Varid{s},\anonymous ,\Varid{s'})\to \Varid{s}\mathbin{:}\Varid{s'}\mathbin{:}\Varid{ss})\;[\mskip1.5mu \mskip1.5mu]\;\Varid{trans}}$\\
${}$\\
${\Varid{faStates}\mathbin{::}\Conid{Eq}\;\Varid{s}\Rightarrow \Conid{FA}\;\Varid{s}\;\Varid{i}\to [\mskip1.5mu \Varid{s}\mskip1.5mu]}$\\
${\Varid{faStates}\;\Varid{fa}\mathrel{=}\Varid{foldl}\;\Varid{union}\;[\mskip1.5mu \mskip1.5mu]\;[\mskip1.5mu \Varid{faStarts}\;\Varid{fa},}$\\
${\phantom{\Varid{faStates}\;\Varid{fa}\mathrel{=}\Varid{foldl}\;\Varid{union}\;[\mskip1.5mu \mskip1.5mu]\;[\mskip1.5mu \mbox{}}\Varid{transStates}\mathbin{\$}\Varid{faTrans}\;\Varid{fa},}$\\
${\phantom{\Varid{faStates}\;\Varid{fa}\mathrel{=}\Varid{foldl}\;\Varid{union}\;[\mskip1.5mu \mskip1.5mu]\;[\mskip1.5mu \mbox{}}\Varid{faFinals}\;\Varid{fa}\mskip1.5mu]}$
\end{tabbing}

We also require the following three functions: the first two implement
Lemmas~\ref{subset-construction} and \ref{intersection-construction},
and the third computes the disjoint union of a list of automata.  To
expedite our exposition, we provide only their type signatures:


\begin{tabbing}
\qquad\=\hspace{\lwidth}\=\hspace{\cwidth}\=\+\kill
${\Varid{faDet}\mathbin{::}(\Conid{Ord}\;\Varid{s},\Conid{Eq}\;\Varid{i})\Rightarrow \Conid{FA}\;\Varid{s}\;\Varid{i}\to \Conid{FA}\;[\mskip1.5mu \Varid{s}\mskip1.5mu]\;\Varid{i}}$\\
${\Varid{faIntersect}\mathbin{::}(\Conid{Eq}\;\Varid{s},\Conid{Eq}\;\Varid{s'},\Conid{Eq}\;\Varid{i})\Rightarrow }$\\
${\hskip9.00em\relax\Conid{FA}\;\Varid{s}\;\Varid{i}\to \Conid{FA}\;\Varid{s'}\;\Varid{i}\to \Conid{FA}\;(\Varid{s},\Varid{s'})\;\Varid{i}}$\\
${\Varid{faDisjointUnion}\mathbin{::}[\mskip1.5mu \Conid{FA}\;\Varid{s}\;\Varid{i}\mskip1.5mu]\to \Conid{FA}\;(\Conid{Int},\Varid{s})\;\Varid{i}}$
\end{tabbing}\ 

Notice that the first two functions return automata whose states are
represented as lists and pairs of the argument automata's states;
these representations intrinsically encode the Lemmas' canonical
injections.  The function $\Varid{faDisjointUnion}$ returns an
automaton whose states are represented as pairs $(i,s)$ where $s$ is a
state of the $i$th argument automaton $\mc{D}_i$.

We use these representations frequently in the following
implementation of Algorithm~\ref{alg:transducer-building}:

\begin{widetext}
\begin{tabbing}
\qquad\=\hspace{\lwidth}\=\hspace{\cwidth}\=\+\kill
${\Varid{transducerFilterFromDomains}\mathbin{::}(\Conid{Eq}\;\Varid{s},\Conid{Ord}\;\Varid{s},\Conid{Eq}\;\Varid{i})\Rightarrow [\mskip1.5mu \Conid{FA}\;\Varid{s}\;\Varid{i}\mskip1.5mu]\to \Conid{Transducer}\;[\mskip1.5mu (\Conid{Int},\Varid{s})\mskip1.5mu]\;\Varid{i}\;\Conid{Int}}$\\
${\Varid{transducerFilterFromDomains}\;\Varid{faDs}\mathrel{=}}$\\
${\hskip2.00em\relax\Conid{FA}\;(\Varid{faStarts}\;\Varid{faA})\;(\Varid{baseTTrans}\plus \Varid{newTTrans})\;(\Varid{faFinals}\;\Varid{faA})}$\\
${\hskip2.00em\relax\mathbf{where}\;\Varid{faA}\mathrel{=}\Varid{faDet}\mathbin{\$}\Varid{faDisjointUnion}\;\Varid{faDs}\mbox{\qquad-{}-  $\mc{A}:=\Det(\mc{D}_1 \sqcup \cdots \sqcup \mc{D}_n)$}}$\\
${\hskip2.00em\relax\phantom{\mathbf{where}\;\mbox{}}\Varid{baseTTrans}\mathrel{=}[\mskip1.5mu (\Varid{s},(\Varid{a},\Varid{f}\;\Varid{s'}),\Varid{s'})\mid (\Varid{s},\Varid{a},\Varid{s'})\leftarrow \Varid{faTrans}\;\Varid{faA}\mskip1.5mu]}$\\
${\hskip2.00em\relax\phantom{\mathbf{where}\;\mbox{}}\hskip2.00em\relax\mathbf{where}\;\Varid{f}\;\Varid{ss}\mid \Varid{length}\;\Varid{is}\equiv \mathrm{1}\mathrel{=}\Varid{head}\;\Varid{is}\mbox{\qquad-{}-  transition ending in $\mc{D}_i$}}$\\
${\hskip2.00em\relax\phantom{\mathbf{where}\;\mbox{}}\hskip2.00em\relax\phantom{\mathbf{where}\;\Varid{f}\;\Varid{ss}\mbox{}}\mid \Varid{otherwise}\mathrel{=}\mathrm{0}\mbox{\qquad-{}-  synchronization, $\lambda$}}$\\
${\hskip2.00em\relax\phantom{\mathbf{where}\;\mbox{}}\hskip2.00em\relax\phantom{\mathbf{where}\;\Varid{f}\;\Varid{ss}\mbox{}}\mathbf{where}\;\Varid{is}\mathrel{=}\Varid{nub}\mathbin{\$}\Varid{map}\;\Varid{fst}\;\Varid{ss}}$\\
${\hskip2.00em\relax\phantom{\mathbf{where}\;\mbox{}}\Varid{forbiddenPairs}\mathrel{=}[\mskip1.5mu (\Varid{s},\Varid{a})\mid \Varid{s}\leftarrow \Varid{faStates}\;\Varid{faA},\Varid{a}\leftarrow \Varid{faAlphabet}\;\Varid{faA}\mskip1.5mu]\mathbin{\char92 \char92 }[\mskip1.5mu (\Varid{s},\Varid{a})\mid (\Varid{s},\Varid{a},\anonymous )\leftarrow \Varid{faTrans}\;\Varid{faA}\mskip1.5mu]}$\\
${\hskip2.00em\relax\phantom{\mathbf{where}\;\mbox{}}\Varid{newTTrans}\mathrel{=}\Varid{map}\;\Varid{newTransition}\;\Varid{forbiddenPairs}}$\\
${\hskip2.00em\relax\phantom{\mathbf{where}\;\mbox{}}\Varid{newTransition}\;(\Varid{s},\Varid{a})\mathrel{=}(\Varid{s},(\Varid{a},\Varid{o}),\Varid{s'})}$\\
${\hskip2.00em\relax\phantom{\mathbf{where}\;\mbox{}}\hskip2.00em\relax\mathbf{where}\;\Varid{faAsa}\mathrel{=}(\Conid{FA}\;(\Varid{f}\mathbin{:}\Varid{faStates}\;\Varid{faA})\;((\Varid{s},\Varid{a},\Varid{f})\mathbin{:}\Varid{faTrans}\;\Varid{faA})\;[\mskip1.5mu \Varid{f}\mskip1.5mu])\mbox{\qquad-{}-  the automaton $\mc{A}^{s,a}$}}$\\
${\hskip2.00em\relax\phantom{\mathbf{where}\;\mbox{}}\hskip2.00em\relax\phantom{\mathbf{where}\;\mbox{}}\Varid{f}\mathrel{=}[\mskip1.5mu (\mathrm{1}\mathbin{+}\Varid{length}\;\Varid{faDs},\Varid{head}\mathbin{\$}\Varid{faStarts}\mathbin{\$}\Varid{head}\;\Varid{faDs})\mskip1.5mu]\mbox{\qquad-{}-  the fresh state $f$ used to build $\mc{A}^{s,a}$}}$\\
${\hskip2.00em\relax\phantom{\mathbf{where}\;\mbox{}}\hskip2.00em\relax\phantom{\mathbf{where}\;\mbox{}}\Varid{faDetAsaCapA}\mathrel{=}(\Varid{faDet}\;\Varid{faAsa})\mathbin{`\Varid{faIntersect}`}\Varid{faA}\mbox{\qquad-{}-  the automaton $\Det(\mc{A}^{s,a} \cap \mc{A})$}}$\\
${\hskip2.00em\relax\phantom{\mathbf{where}\;\mbox{}}\hskip2.00em\relax\phantom{\mathbf{where}\;\mbox{}}\Varid{faZ}\mathrel{=}\Varid{faDet}\mathbin{\$}\Varid{zero}\;\Varid{faDetAsaCapA}\mbox{\qquad-{}-  the automaton $\Det(\mc{Z}[\Det(\mc{A}^{s,a} \cap \mc{A})])$}}$\\
${\hskip2.00em\relax\phantom{\mathbf{where}\;\mbox{}}\hskip2.00em\relax\phantom{\mathbf{where}\;\mbox{}}\Varid{reachableStateSeq}\mathrel{=}\Varid{take}\;(\Varid{length}\mathbin{\$}\Varid{faStates}\;\Varid{faZ})\mbox{\qquad-{}-  the states $s_1, \dots, s_{m+m'}$}}$\\
${\hskip2.00em\relax\phantom{\mathbf{where}\;\mbox{}}\hskip2.00em\relax\phantom{\mathbf{where}\;\mbox{}}\phantom{\Varid{reachableStateSeq}\mathrel{=}\mbox{}}(\Varid{iterate}\;\Varid{nextState}\mathbin{\$}\Varid{head}\mathbin{\$}\Varid{faStarts}\;\Varid{faZ})}$\\
${\hskip2.00em\relax\phantom{\mathbf{where}\;\mbox{}}\hskip2.00em\relax\phantom{\mathbf{where}\;\mbox{}}\hskip2.00em\relax\mathbf{where}\;\Varid{nextState}\;\Varid{s}\mathrel{=}\Varid{head}\;[\mskip1.5mu \Varid{s''}\mid (\Varid{s'},\anonymous ,\Varid{s''})\leftarrow \Varid{faTrans}\;\Varid{faZ},\Varid{s'}\equiv \Varid{s}\mskip1.5mu]}$\\
${\hskip2.00em\relax\phantom{\mathbf{where}\;\mbox{}}\hskip2.00em\relax\phantom{\mathbf{where}\;\mbox{}}\Varid{sStarLs}\mathrel{=}\Varid{map}\;((\Varid{map}\;\Varid{snd})\mathbin{\circ}\mbox{\qquad-{}-  the sets $\{S_{*,l}\}_{l=1}^{m+m'}$}}$\\
${\hskip2.00em\relax\phantom{\mathbf{where}\;\mbox{}}\hskip2.00em\relax\phantom{\mathbf{where}\;\mbox{}}\phantom{\Varid{sStarLs}\mathrel{=}\Varid{map}\;(\mbox{}}(\Varid{intersect}\mathbin{\$}\Varid{faFinals}\;\Varid{faDetAsaCapA}))\;\Varid{reachableStateSeq}}$\\
${\hskip2.00em\relax\phantom{\mathbf{where}\;\mbox{}}\hskip2.00em\relax\phantom{\mathbf{where}\;\mbox{}}\Varid{siStars}\mathrel{=}[\mskip1.5mu \Varid{nub}\mbox{\qquad-{}-  the sets $\{S_{i,*}\}_{i=1}^{}$}}$\\
${\hskip2.00em\relax\phantom{\mathbf{where}\;\mbox{}}\hskip2.00em\relax\phantom{\mathbf{where}\;\mbox{}}\phantom{\Varid{siStars}\mathrel{=}[\mskip1.5mu \mbox{}}[\mskip1.5mu \Varid{s}\mid \Varid{s}\leftarrow \Varid{map}\;\Varid{snd}\mathbin{\$}\Varid{faFinals}\;\Varid{faDetAsaCapA},\Varid{length}\;\Varid{s}\equiv \Varid{i}\mskip1.5mu]}$\\
${\hskip2.00em\relax\phantom{\mathbf{where}\;\mbox{}}\hskip2.00em\relax\phantom{\mathbf{where}\;\mbox{}}\phantom{\Varid{siStars}\mathrel{=}\mbox{}}\mid \Varid{i}\leftarrow [\mskip1.5mu \mathrm{1}\mathinner{\ldotp\ldotp}\Varid{length}\mathbin{\$}\Varid{head}\mathbin{\$}\Varid{faStarts}\;\Varid{faA}\mskip1.5mu]\mskip1.5mu]}$\\
${\hskip2.00em\relax\phantom{\mathbf{where}\;\mbox{}}\hskip2.00em\relax\phantom{\mathbf{where}\;\mbox{}}\Varid{sijs}\mathrel{=}\Varid{concatMap}\;(\lambda \Varid{siStar}\to \Varid{map}\;(\Varid{intersect}\;\Varid{siStar})\;\Varid{sStarLs})\;\Varid{siStars}\mbox{\qquad-{}-  the sets $\{S_{i,j}\}_{i,j}$}}$\\
${\hskip2.00em\relax\phantom{\mathbf{where}\;\mbox{}}\hskip2.00em\relax\phantom{\mathbf{where}\;\mbox{}}[\mskip1.5mu \Varid{s'}\mskip1.5mu]\mathrel{=}\Varid{head}\mathbin{\$}\Varid{filter}\;(\lambda \Varid{sij}\to \Varid{length}\;\Varid{sij}\equiv \mathrm{1})\;\Varid{sijs}\mbox{\qquad-{}-  the state $s'$ to which to synchronize}}$\\
${\hskip2.00em\relax\phantom{\mathbf{where}\;\mbox{}}\hskip2.00em\relax\phantom{\mathbf{where}\;\mbox{}}\Varid{o}\mathrel{=}\mathbin{-}\mathrm{1}\mbox{\qquad-{}-  domain break}}$\\
${\hskip2.00em\relax\phantom{\mathbf{where}\;\mbox{}}\Varid{zero}\;\Varid{fa}\mathrel{=}\Conid{FA}\;(\Varid{faStarts}\;\Varid{fa})\;[\mskip1.5mu (\Varid{s},\mathrm{0},\Varid{s'})\mid (\Varid{s},\anonymous ,\Varid{s'})\leftarrow \Varid{faTrans}\;\Varid{fa}\mskip1.5mu]\;(\Varid{faFinals}\;\Varid{fa})}$
\end{tabbing}

\end{widetext}

To implement the domain optimization algorithm
$\mrm{Optimize}(\bullet)$ is somewhat more---but not
overwhelmingly---complicated.  In fact, we generated the examples
here by computer, rather than by hand.

\bibliographystyle{unsrt}
\bibliography{apd.refs}

\end{document}